\documentclass[journal, english]{IEEEtran}

\ifCLASSINFOpdf
            \else
                  \fi

\usepackage{pgfplots} \newlength\fheight \newlength\fwidth \newlength{\figureheight}
\newlength{\figurewidth}
\usepackage{tikz}

\pgfplotsset{compat=1.9}

\tikzset{pics/myrec/.style n args={3}{code={\draw[very thick, #3] (0,0) rectangle (#1,#2); 
}}, 
pics/myrec/.default={1}{0}{pink},
} 

\usepgfplotslibrary{groupplots}

\usepackage{textcomp}

\usetikzlibrary{decorations.pathreplacing}
\usetikzlibrary{shapes.geometric,arrows}

\usepackage{subfigure} 
\usepackage{caption}

\usepackage{import}

\usepackage{booktabs}
\usepackage{diagbox}

\usepackage{rotating}

\usepackage{float}
\usepackage{xcolor}

\usepackage{babel}
\usepackage{nohyperref}
\usepackage{url}

\usepackage{tikz-3dplot}

\usepackage{array}
\usepackage{multirow}

\usepackage[acronym]{glossaries}

\usepackage{cite}

\newcommand{\ingreen}[1]{{\color{black}  #1}}
\newcommand{\inblue}[1]{{\color{black} #1}}

\newcommand\MyBox[1]{
  \fbox{\lower0.75cm
    \vbox to 0.8cm{\vfil
      \hbox to 1.2cm{\hfil\parbox{1.6cm}{#1}\hfil}
      \vfil}  }}

\usetikzlibrary{positioning}

\DeclareMathSizes{10}{10}{5}{5}

\usepackage{siunitx}

\usepackage{makecell}

\usepackage{luatex85,shellesc}
\usepackage{balance}

\usepackage{amsmath}

\usepackage{algpseudocode}
\usepackage[ruled,vlined]{algorithm2e}
\usetikzlibrary{spy}
\usetikzlibrary{bayesnet}
\usepackage[utf8]{inputenc}
\usepackage{booktabs,ragged2e}

\usepackage[flushleft]{threeparttable}

\usetikzlibrary{backgrounds}

\usepackage{pdfpages}
\usepackage[skins]{tcolorbox}

\usepackage{amssymb}

\DeclareMathOperator*{\argmin}{arg\,min}

\usepackage{mathtools}

\usepackage[mode=buildnew]{standalone}

\newacronym{ADAS}{ADAS}{automatic driver assistance system}

\newacronym{CNN}{CNN}{convolutional neural network}
\newacronym{COOS}{COOS}{coordinate system}

\newacronym{DRISF}{DRISF}{deep rigid instance scene flow}
\newacronym{DRISFwR}{DRISFwR}{deep rigid instance scene flow with radar}
\newacronym{DOA}{DoA}{direction-of-arrival}
\newacronym{DGPS}{DGPS}{Differential-GPS}
\newacronym{DGPS-INS}{DGPS-INS}{Differential-GPS with inertial navigation system}
\newacronym{DNN}{DNN}{deep neural network}

\newacronym{FMCW}{FMCW}{frequency modulated continuous wave}
\newacronym{FOV}{FoV}{field of view}

\newacronym{GN}{GN}{Gau{\ss}-Newton}

\newacronym{MAE}{MAE}{mean absolute error}
\newacronym{MTI}{MTI}{moving target indication}

\newacronym{NN}{NN}{neural network}

\newacronym{RD}{RD}{range-doppler}
\newacronym{RGB}{RGB}{Red Green Blue}

\newacronym{CFAR}{CFAR}{Constant false alarm rate}

\newacronym{PDF}{PDF}{probability density function}

\newacronym{BF}{BF}{Bartlett beamforming}
\newacronym{PM}{PM}{phase-comparison monopulse}

\begin{document}
\title{
\fontsize{23}{23} \selectfont
Warping of Radar Data into Camera Image for Cross-Modal Supervision in Automotive Applications 
}

\markboth{This article has been accepted for publication in IEEE Transactions on Vehicular Technology. DOI 10.1109/TVT.2022.3182411}%
{}

%


\makeatletter
\def\ps@IEEEtitlepagestyle{
  \def\@oddfoot{\mycopyrightnotice}
  \def\@evenfoot{}
}
\def\mycopyrightnotice{
  {\footnotesize
  \begin{minipage}{\textwidth}
  \centering
Copyright~\copyright~2015 IEEE. Personal use of this material is permitted. However, permission to use this \\ 
material for any other purposes must be obtained from the IEEE by sending a request to pubs-permissions@ieee.org. 
  \end{minipage}
  }
}

\author{\mbox{Christopher~Grimm}, \IEEEmembership{Graduate Student Member, IEEE},
\mbox{Tai~Fei}, \IEEEmembership{Senior Member, IEEE},
\mbox{Ernst~Warsitz},
\mbox{Ridha Farhoud},
\mbox{Tobias Breddermann},
\mbox{Reinhold~Haeb-Umbach, \IEEEmembership{Fellow, IEEE}}

\thanks{Christopher~Grimm,~Tai~Fei,~Ernst~Warsitz,~Ridha Farhoud and \mbox{Tobias Breddermann} are employees of \mbox{HELLA GmbH \& Co. KGaA}, 59552 Lippstadt, Germany e-mail: \{christopher.grimm, tai.fei, ernst.warsitz, ridha.farhoud, tobias.breddermann\}@forvia.com}\thanks{Reinhold~Haeb-Umbach is with the Department of Communication Engineering, Paderborn University, 33098 Paderborn, Germany e-mail: haeb@nt.uni-paderborn.de}}


\maketitle

\begin{abstract}

We present an approach to automatically generate semantic labels for real recordings of automotive range-Doppler (RD) radar spectra. Such labels are required when training a neural network for object recognition from radar data. The automatic labeling approach rests on the simultaneous recording of camera and lidar data in addition to the radar spectrum.
By  warping radar spectra into the camera image, state-of-the-art object recognition algorithms can be applied to label relevant objects, such as cars, in the camera image. 
The warping operation is designed to be fully differentiable, which allows backpropagating the gradient computed on the camera image through the warping operation to the neural network operating on the radar data.  
As the warping operation relies on accurate scene flow estimation, we further propose a novel scene flow estimation algorithm which exploits information from camera, lidar and radar sensors. The proposed scene flow estimation approach is compared against a state-of-the-art scene flow algorithm, and it outperforms it by approximately 30\% w.r.t. mean average error.
The feasibility of the overall framework for automatic label generation for RD spectra is verified by evaluating the performance of neural networks trained with the proposed framework for Direction-of-Arrival estimation.

\end{abstract}

\vspace{0.2cm}

\IEEEpeerreviewmaketitle

\newlength\figH
\newlength\figW

\section{Introduction}\label{sec:Introduction}

In the past decade, \gls{DNN} based algorithms have proven excellent performance for a wide range of applications  \cite{Haeb-Umbach_Heymann_Drude_Watanabe_Delcroix_Nakatani_2021, Cir12, zhu17}. Those impressive results rely, however, to a great degree on the availability of large labeled databases. In case of \gls{RD} spectrum automotive \gls{FMCW} radar signal processing this requirement poses a significant practical problem. Not only are those large-scale databases not available as of today, they would be furthermore very difficult to build, because, unlike, e.g., a camera image, an \gls{RD} plot is very difficult for a human to interpret and thus to label properly. \inblue{Another challenge comes from high variations in radar hardware and parameterization, resulting in very specific radar measurement capabilities. Therefore, researchers often require to generate their own dataset, rather than using open public datasets  \cite{zhang2021raddet}.}

Some recent works have dealt with automatic \gls{RD} spectrum labeling by lidar and/or camera \cite{Schult18, Maj19, Lim2019RadarAC, ouaknine2020carrada}. However, these labels are generally restricted to object level for certain classes as pedestrian and vehicles, which limits its use to ``high-level'' radar classification tasks, such as moving object detection. We, on the contrary, are here interested in applying \glspl{DNN} to low-level radar classification tasks, such as radar target detection, stationary target detection and \glspl{DOA} estimation.

In this paper, we address this problem by automatically generating high-quality and dense labels from a reference system consisting of camera, lidar and state-of-the-art visual object classification algorithms. Neural object classification algorithms are applied to the camera image to label relevant objects. 
Those labels are then mapped to locations in an \gls{RD} spectrum by warping the \gls{RD} spectrum into the camera image, and then serve as labels for a neural network operating on radar spectrum data. The warping task, however, is challenging,  as radar and camera measure quite different physical phenomena. Furthermore, the field of view of camera and radar may only partially overlap, and their coordinate systems are not aligned.  

We augment the measurement space of the reference camera system to cover the measurement space of the radar system by estimating the 3D-velocity of each pixel in the  camera image. To this end, we extend a previously proposed scene flow estimation algorithm to utilize information gained from the radar spectrum.  This significantly improves scene flow estimation, and in turn the precision of the warping operation.

\begin{figure}[h]
  \centering
{
\hspace{-0.7cm}\includestandalone[width=0.5\textwidth]{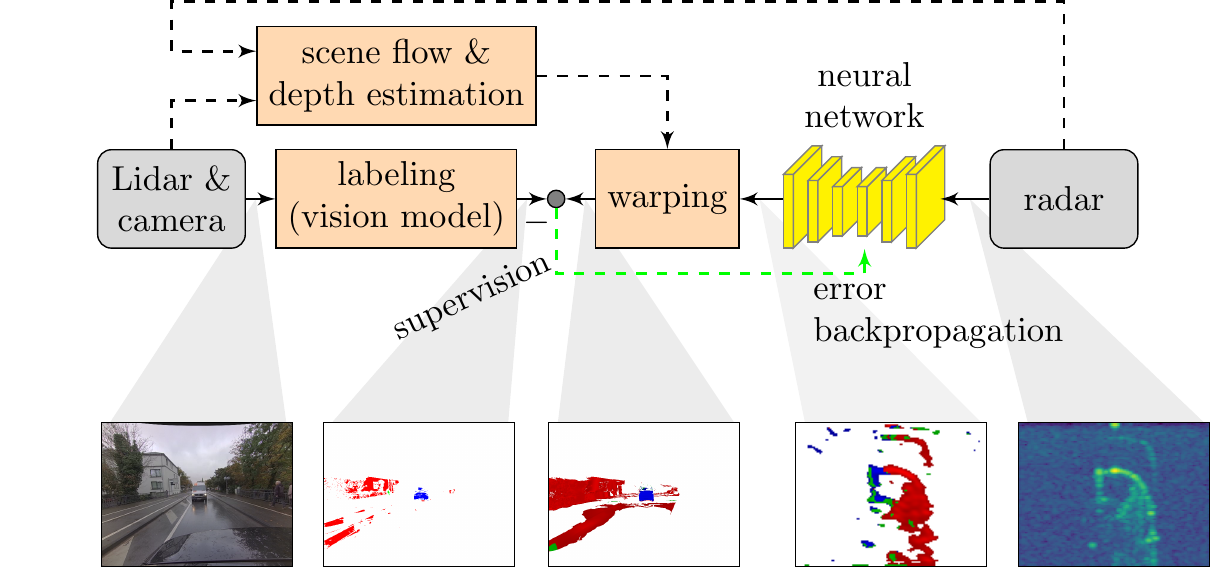}
  }
\caption{\textbf{Overview of the proposed approach:}
Lidar and camera provide sensor readings to the vision model (typically a \gls{DNN}) which automatically generates labels in the camera image. At the same time, the radar's RD spectrum is processed by a neural network and warped from \gls{RD} spectrum into the camera image domain. The residual between the generated label and the warped neural network prediction is calculated and then backpropagated through the warping operation to the network operating on the radar data for parameter adjustment (green arrow). The warping operation requires dense scene flow and depth estimation, which is obtained by lidar, camera and radar data.
The lower row shows example  images of a semantic segmentation application. From left to right: RGB image, reference semantic mask (only relevant pixels shown), predicted semantic mask warped into camera image, predicted semantic mask in RD-spectrum, RD-map.}
    \label{fig:ablaufdiagram3}
\end{figure}

As the warping operation is designed to be fully differentiable, the gradient computed on the camera image can be backpropagated through the warping module to the neural network operating  on the RD spectrum, allowing for supervised training of the network with real world radar data as input.
Fig. \ref{fig:ablaufdiagram3} illustrates the proposed framework. It can be viewed as a universal framework for cross-modal supervision of neural networks operating on \gls{FMCW} radar data.

We demonstrate the effectiveness of the proposed framework by applying it to radar based \gls{DOA} estimation and present  both qualitative and quantitative results. 

The main contributions of this paper can be summarized as follows
\begin{itemize}
\item An improved camera scene flow estimation algorithm by incorporating RD spectrum information
\item Dense label generation for RD spectrum of FMCW radar data by utilizing labels from camera vision models
\item A method for warping radar data into camera images
\item A framework to enable supervised training of \glspl{DNN} on real-world data for multiple radar applications.
\end{itemize}

The remainder of this paper is organized as follows. \mbox{Sec. \ref{se:rw}} summarizes related work in the field of automotive dataset generation. Sec. \ref{se:cs} introduces the  utilized sensors and the basic sensor calibration procedure. Sec. \ref{ss:sceneflow} describes the novel scene flow estimation algorithm which is supported by radar measurements. The scene flow estimates are used in the warping algorithm given in  Sec. \ref{ss:warping}. In Sec. \ref{ss:appliedSupervision}, camera labels, training objectives for the \glspl{NN} and network architectures are given as well as techniques on how to make the training more robust against reference label errors. In Sec. \ref{ss:nnEvaluation} the NN based DoA estimator's accuracy is evaluated. The conclusions are drawn in Sec. \ref{ss:outlook}.

\section{Related work}\label{se:rw}

In 2012, Geiger \textit{et al.} \cite{Geiger12b} introduced the famous \textit{KITTI} dataset. The development and publication of this dataset was motivated by the intention to foster the development of computer vision for autonomous driving. At that time,  algorithms were  mostly evaluated under laboratory conditions and performed poorly when  applied to field  data \cite{Geiger12c}. Recognizing the need for more realistic datasets, the authors collected camera, lidar and GPS data from approx. $\SI{40}{\kilo\meter}$  real-world driving scenarios around Karlsruhe, Germany. Starting with benchmarks for stereo vision, optical flow, visual odometry and 3D object detection, in which ground truth was provided either by sensor fusion or, in case of object detection, by human labelers, the range of applications and benchmark systems has been extended since then to depth completion of sparse depth measurements, semantic segmentation and scene flow estimation \cite{Menze2015CVPR}. However, automotive radar has not been included in the dataset, and, accordingly, \textit{KITTI} is not suitable for the development of automotive radar applications considered here. 

In 2019, Caeser \textit{et al.} \cite{Caesar19} introduced the \textit{nuScenes} multimodal dataset, which incorporates sensor data from multiple cameras, lidar and radars. The ground truth was mostly provided by professional human labelers, for instance in the form of 3D object bounding boxes. The dataset currently contains approx. $\SI{242}{\kilo\meter}$ of driving data acquired in Boston, USA, and in Singapore. The radar data is given as a point cloud consisting of radar detections, in which each point has an annotated position and velocity. From the perspective of radar signal processing, the radar data in this dataset is already above spectrum level, while raw spectrum level data is missing. Accordingly, it is barely suitable for the development of signal processing algorithms for RD spectrum data.

Similar concerns apply to the dataset provided in \cite{Meyer2019AutomotiveRD}. Other  automotive datasets can be found e.g., in \cite{Cordts16, Neuhold2017TheMV} which, however, are concerned with creating large and densely annotated camera data only.

Recently, more and more synthetic datasets have been developed.  The datasets of  \cite{Ros16, Gaidon16, Dosovitskiy17} contain artificially generated camera images as well as detailed ground truth from simulated environments. Bühren \textit{et al.} \cite{Buehren06}  synthesized radar targets from traffic simulations. Again, the simulation was carried out on above spectrum level radar data, and thus is  inappropriate for the development \inblue{of} signal processing algorithms for RD spectra.

In \cite{Sli20} the authors showcased deep-learning based object detection trained on simulated RD spectra via raytracing. Different driving scenarios were simulated and a trained object detector was able to achieve decent performance on the test data. Note, however, that the test data originated from the same raytracing tool as the training data, and thus performance on real world data remains unknown. 

In \cite{Shuqing10} the authors propose a \gls{DGPS} based reference system for radar target evaluation. While this provides very precise measurements, the equipment is very expensive and, thus, the acquisition of large and complex datasets is practically impossible as only objects are covered, equipped with such a system. Reference RD spectra can also be generated from lidar data, as proposed in \cite{Schult18}. Vehicles are detected in the lidars point cloud and tracked over time. However, the reference RD spectra are very sparse, since only a reference for vehicles was given, and since vehicles are modeled as single point reflectors.

The idea of cross-modal supervision was followed in a number of works.
Zhao \textit{et al.}  proposed a framework to leverage camera based label predictions for other sensors\cite{Zhao18}.
In 2019, \cite{Maj19} and \cite{Lim2019RadarAC} trained \glspl{DNN} for object detection operating on range-azimuth-Doppler spectrum in real world driving scenarios. The radar data is mapped from polar coordinates to the Cartesian coordinate system of the reference lidar sensor, in which the object labels are presented to the network. However, the Cartesian coordinate transformation  relies on accurate azimuth resolution of the radar itself, which can hardly be met by today's series production radar sensors for \gls{ADAS} applications. Additionally, the framework was designed to only generate labels for vehicles on highway environments and thus only provides a sparse label subset of the entire scene.

\begin{table*}[h]
\centering
\caption{Comparison of automatic labeling frameworks. 
}
\label{tab:FramworkComparison}
\begin{tabular*}{1.0\textwidth}{lccc}
\toprule
Framework & \cite{Maj19} / \cite{Lim2019RadarAC} & \inblue{\cite{zhang2021raddet}} / \cite{ouaknine2020carrada}& ours\\
\midrule
Sensors (lidar/camera/radar)& yes/no/yes & no/yes/yes &yes/yes/yes\\
Sceneflow approach & None & Optical flow based / no & State-of-the-art \\
Classes (Pedestrian, car, stationary, background) & yes, yes, no, yes & yes, yes, no, yes & yes, yes, yes, yes \\
Label density & sparse & sparse & dense \\
Scene complexity in dataset & highway & test track / urban + highway & urban + highway\\
Applications \\
(Semantic segmentation, DoA,Target detection, RCS estimation from camera)  & yes, no, no, no & yes, no, no, no & yes, yes, yes, yes\\
\bottomrule
\end{tabular*}
\end{table*}

In \cite{Wang19, Wang19b} the authors labeled radar target clusters by fusing camera images, lidar pings and radar targets with the help of state-of-the-art perception models and achieved \mbox{72 \%} classification accuracy compared to human labeling with a significant reduction in labeling time. They achieved $\SI{0.48}{\meter}$ average translational error on the \textit{nuScenes} object detection benchmark. But, again, this task is above spectrum level, and, therefore,  not applicable here.

One of the very few sets of annotated radar spectrum data is \cite{ouaknine2020carrada}, in which object annotations for vehicles, cyclists and pedestrians are given in range-azimuth-Doppler space. The annotations are generated by assigning object classes from objects estimated by camera to matching objects estimated by radar. Although annotations are given for spectrum data, the accuracy of the automatically generated annotations highly depends on the accuracy of the radar based object detections. Furthermore, the matching becomes significantly harder in dense urban environments with multiple objects per scene, potentially resulting in a high number of mismatches and therefore introducing label noise. The framework does not provide labels for stationary environment and thus provides only a  subset of all labels in a complex environment.

\inblue{
In 2021, a dataset containing Range-Azimuth-Doppler spectra from radar was presented in \cite{zhang2021raddet}. Object level labels are provided for the classes: person, bicycle, car, motorcycle, bus and truck. To auto-generate the labels, radar detection pointclouds are extracted, by applying CFAR and beamforming to the radar spectra. The pointcloud is clustered into radar objects, which are then associated with class labels from instance segmentation on stereo image pairs. This approach is very similar to the approach presented in \cite{ouaknine2020carrada} and therefore we see similar restrictions: \ingreen{Firstly, the} label quality highly relies on the quality of the radar pointcloud. The second issue is that the labels provided here are limited to dynamic road users only. Adding labels for stationary objects would likely compromise the association accuracy between radar and camera objects. 
}

Based on our research, we ranked the best suitable frameworks for radar RD spectrum labeling from literature against our proposed framework in \mbox{Tab. \mbox{\ref{tab:FramworkComparison}}}. We combined the approaches from \cite{Maj19} and \cite{Lim2019RadarAC} as well as \cite{ouaknine2020carrada} and \cite{zhang2021raddet} as they present very similar approaches for auto-annotation. As prior publications focus on providing annotations for the application of semantic segmentation in radar spectra, we do see significant difference to our dataset, which is also able to provide dense labels for DoA estimation, as proven in this paper and other applications.




\inblue{
The reader will find information about classical DoA radar applications e.g. phase-monopulse or Bartlett beamforming in e.g. \cite{Stabilito61}, \cite{Krim96}. 

In 2012 Sit \textit{et al.} \cite{Sit2012NeuralNB} proposed a NN based DoA estimator. The NN was trained 859 samples referring two scatters moving in azimuth. The NN was able to achieve better DoA separability than classical MUSIC algorithm whilst offering significant reduced processing time.

More recent NN based approaches were presented in \cite{Gar18, Fuc19a, Fuc19b}. The authors installed a radar-under-test on an azimuth positioner in an anechoic chamber and placed a corner reflector \SI{1.5}{\meter} in front of the sensor. Measurement data from the radar sensor was acquired at different azimuth orientations. As the target position was known, it served as ground truth label for multi-layer perceptron (MLP) training. The MLP was trained to estimate DoA for up to two targets, and achieved similar performance as classical algorithms for instance Deterministic Maximum Likelihood, while being significantly faster in inference. As the data collected  in anechoic chamber might differ from data obtained in real world traffic scenarios, we will focus on training and evaluating NNs in the latter scenarios.

For the interested reader, a more elaborated overview can be found in \cite{You20}.

}


\inblue{
\section{Sensor calibration}\label{se:cs}
}

In the following we assume the presence of the following sensors, which reflects our measurement setup  described in Section~\ref{ss:measurement_setup} further below:
\begin{itemize}
	\item $D$: A \gls{DGPS-INS} which delivers high-precision position information
	\item $C_1, C_2$: An array of two cameras, one pointing to the rear, and the other  to the right.
	\item $L_1, L_2$: Two lidar scanners
	\item $R$: the RD spectrum radar unit.
\end{itemize}
The spatial information they provide is given in their local \glspl{COOS}. Since the sensors are not mounted at the same position, the origins of their local \glspl{COOS} differ.  
For jointly processing the data, the \glspl{COOS} of the sensors and of the (ego) vehicle, denoted by $E$, have to be aligned first.



A transformation of a position $\mathbf{x}_j$ in the \gls{COOS} of sensor $j$ to the position $\mathbf{x}_i$ in the \gls{COOS} of sensor $i$, where $i,j \in \{D, C_1, C_2, L_1, L_2, R, E\}$, is achieved by \inblue{affine transformation} with the rotation matrix ${}^{i}\textbf{R}_\text{j}\in \rm I\!R^{3\times3}$ and the translation vector ${}^{i}\textbf{t}_\text{j}\in \rm I\!R^{3\times1}$ as follows:
\begin{align}
\label{eq:coos_transformation}
	\mathbf{x}_i &= {}^{i}\textbf{R}_\text{j} \mathbf{x}_j + {}^{i}\textbf{t}_\text{j}.
\end{align}
The parameters $({}^{E}\textbf{R}_\text{L1}, {}^{E}\textbf{t}_\text{L1})$ and \textbf{$({}^{E}\textbf{R}_\text{L2}, {}^{E}\textbf{t}_\text{L2})$}
from the \gls{COOS} of the lidar sensors $L_1$ and $L_2$ to the \gls{COOS} of the ego vehicle have been estimated via \cite{Thrun14}. The tuples $({}^{C1}\textbf{R}_\text{L1}, {}^{C1}\textbf{t}_\text{L1})$ and $({}^{C2}\textbf{R}_\text{L1}, {}^{C2}\textbf{t}_\text{L1})$ have been extracted from sensor mounting parameters, because the cameras were mounted right next to the lidar sensors, see Fig.~\ref{fig:vehicleConfiguration} in Section~\ref{ss:nnEvaluation}. The radar position relative to the ego vehicle $({}^{R}\textbf{R}_\text{E}, {}^{R}\textbf{t}_\text{E})$ was measured via \cite{kuehnau17}, including intrinsic calibration. The DGPS-INS position $({}^{E}\textbf{R}_\text{D}, {}^{E}\textbf{t}_\text{D})$ has been measured via measuring tape. Transformation between other \gls{COOS} pairs are derived from these tuples by subsequent transformation, e.g. the transformation from ego to camera \gls{COOS} $({}^{C}\textbf{R}_\text{E}, {}^{C}\textbf{t}_\text{E})$ can be derived by transformation from ego to lidar first and then from lidar to camera.

The output of the camera is an image. Angular information is coded in the image pixels $\textbf{p}_i$ for camera \mbox{$i\in\{1,2\}$}. This information has to be first converted to camera coordinates $\textbf{x}_{C1}$ and $\textbf{x}_{C2}$. This is done by applying a camera-pinhole model \cite{xu96} from Eq. \ref{eq:coosTrafo_e} and \ref{eq:coosTrafo_f} with the intrinsic parameters focal lengths $\{f_{x1}, f_{y1}, f_{x2}, f_{y2}\}$ and principal points $\{c_{x1}, c_{y1}, c_{x2}, c_{y2}\}$, estimated via \cite{Geiger12b}
\begin{IEEEeqnarray}{rCl}
\textbf{p}_{i} &=& [u_{i}, v_{i}]^T\IEEEyesnumber\IEEEyessubnumber\label{eq:coosTrafo_g}\\
x_{Ci,x} &=& \frac{u_{i} - c_{xi}}{f_{xi}}x_{Ci,z} \IEEEyessubnumber\label{eq:coosTrafo_e}\\
x_{Ci,y} &=& \frac{v_{i} - c_{yi}}{f_{yi}}x_{Ci,z} \IEEEyessubnumber\label{eq:coosTrafo_f},
\end{IEEEeqnarray}
where $\textbf{x}_{Ci} = (x_{Ci,x}, x_{Ci,y}, x_{Ci,z})^T$.
As the remainder of this paper applies to both cameras,  we will from now on discard the camera index $i$ for the sake of simplicity. The term $x_{Ci,z}$ is known by dense depth estimation, which will be discussed later.

\section{Novel scene flow estimation approach}\label{ss:sceneflow}
In order to warp the radar RD data into camera image, which is part of the overall processing chain shown in Fig. \ref{fig:ablaufdiagram3}, we first have to estimate the 3-D velocity of every pixel in the camera image. This process is commonly referred as scene flow estimation. 

In \cite{Ma19}, a scene flow estimation algorithm called \gls{DRISF}  was proposed, which achieves state-of-the-art performance on the \textit{KITTI} data set at the time of the development of this paper. The idea is, that the overall scene flow can be estimated by considering the 3-D motion of each actor individually. \Gls{DRISF} is based on the definition of an objective function consisting  of energy terms, that is optimized by a \gls{GN} approach.

\begin{figure}[h]
  \centering
\includestandalone[width=\columnwidth]{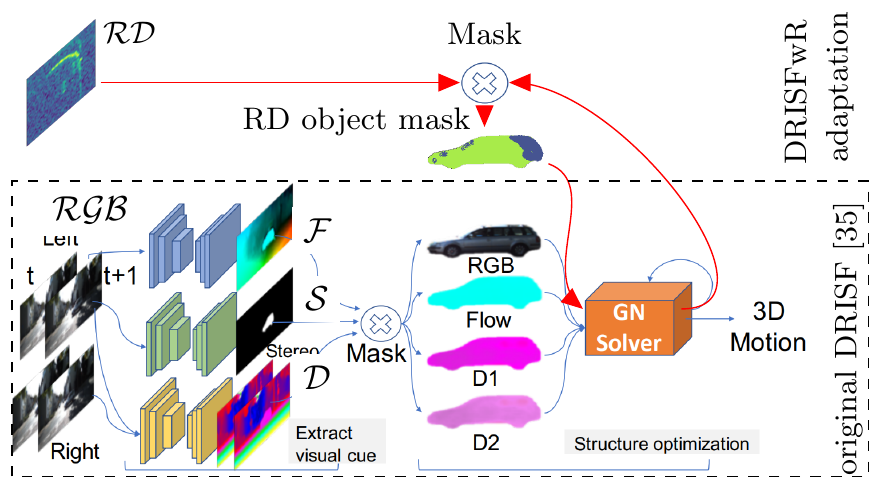}
\caption{\textbf{DRISFwR overview (adapted from \cite{Ma19}):} The original DRISF approach is drawn in the dashed rectangle. Our DRISFwR extension is shown on top. After each iteration of the \gls{GN} solver, the RD object mask is updated and used for next iteration.}
    
    \label{fig:drisfwrOververview}
\end{figure}

As we found the accuracy of DRISF insufficient to achieve satisfactory warping performance, we here propose an extension called \gls{DRISFwR}, which adds an energy term computed from RD spectrum data to the objective function of DRISF. This radar specific adaptation is schematically depicted in \mbox{Fig. \ref{fig:drisfwrOververview}}.

\Gls{DRISFwR} not only reduces scene flow errors, as will be illustrated in Sec.~\ref{ss:doaEval}, it also integrates automatic alignment of scene flow estimation with radar data. 
Furthermore, we incorporate precise ego-motion from DGPS-INS into the algorithm to achieve an accurate estimation of static background scene flow.

In the following, we first describe the input features for \gls{DRISFwR}, then define required pixel subsets, explain the process of static background scene flow estimation and the required adaptation of the energy formulation for foreground objects.

\subsection{Input data to DRISFwR}
Four types of knowledge sources are exploited for scene flow estimation: instance segmentation, optical flow, dense depth information and RD spectrum. They are depicted in Fig. \ref{fig:drisfwrOververview} as $\mathcal{F}$, $\mathcal{S}$, $\mathcal{D}$ and $\mathcal{RD}$.

\subsubsection*{\textbf{Optical Flow mask}} For optical flow calculation between two adjacent RGB frames, $\text{HD}^{3}$-Flow \cite{Yin19} has been applied. It learns probabilistic pixel correspondences to provide optical flow confidence at pixel level and achieves state-of-the-art results on the  \textit{KITTI} dataset. In the following, let $\mathcal{F}$ denote the computed flow masks, which contain the optical flow values for all the pixels in the camera image $\mathcal{RGB}$.

\subsubsection*{\textbf{Instance Segmentation mask}} Segmentation information of relevant objects in the camera's image is obtained by applying a pretrained \gls{NN} for object segmentation to the camera image. 
Here, semantic instance segmentation is produced via mask\_rcnn\_inception\_v2\_coco from the Tensorflow model zoo \cite{tfZoo20}. This NN was trained on the COCO dataset and  selected here for its good accuracy and fast inference time. During inference, semantic object masks $\mathcal{S}$ are generated as well as object classes. Since the utilized network is not specifically tailored for automotive classes, we retain only the following five relevant classes: pedestrian, car, truck, bicycle and motor bike.

\subsubsection*{\textbf{Dense depth mask}} 
\begin{figure}[h]
  \centering
{
\includestandalone[width=0.5\textwidth]{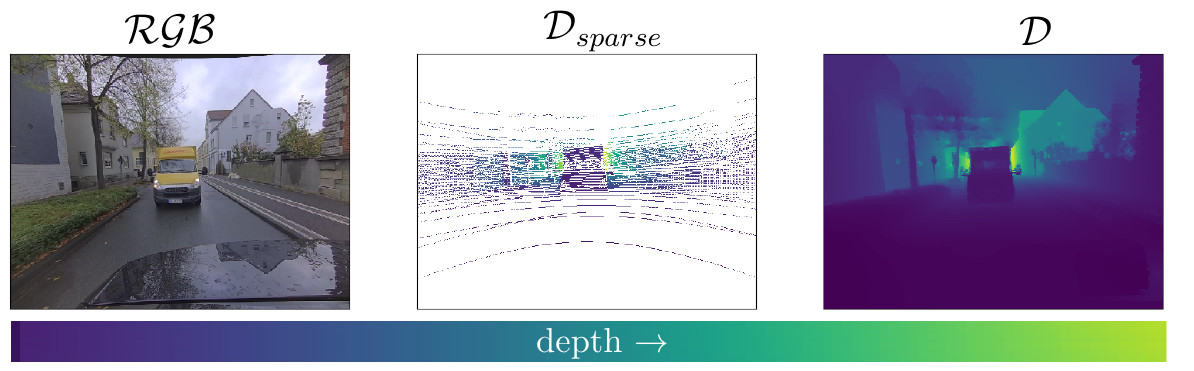}
  }
\vspace{-0.6cm}
\caption{\textbf{Depth completion results:} Left-to-right: RGB image, sparse depth information for those pixels that correspond to lidar measurements, dense depth obtained by depth completion.}
    \label{fig: depthCompletion}
\end{figure}

In DRISF \cite{Ma19}, Ma \textit{et al.} obtained depth estimation based on the stereo camera setup of the \textit{KITTI} dataset. As our sensor setup does not have a stereo camera setup, we  utilize sparse depth measurements \mbox{$\mathcal{D}_\text{sparse}\coloneqq \{ \textbf{p} \mid \textbf{p}\in \text{lidar}\}$} of those camera pixels that are in the lidars field of view, and ``densify'' them  with the help of  mono-cameras. This process is called depth completion with lots of research done in \cite{Geiger12b}, \cite{Diebel06}. We employed the algorithm proposed by \cite{Harrison09} which solves depth completion via Markov Random Fields and which requires no adjustment to our sensor setup. The algorithm projects the sparse lidar depth measurements onto the camera image using eqs.~\eqref{eq:coos_transformation}, \eqref{eq:coosTrafo_e} and \eqref{eq:coosTrafo_f}, and propagates the depth values from lidar occupied pixels into regions with similar brightness values, while enforcing second-order smoothness in depth, see Fig. \ref{fig: depthCompletion} for an example.

\subsubsection*{\textbf{RD spectrum}}
As a  new cue for scene flow estimation, radar data is employed, which provides valuable velocity measurements from the scene. The non-coherent power spectral density $\mathcal{RD}(\textbf{p}_s)$ of pixel $\textbf{p}_s$ is computed from the complex-valued RD spectrum $\mathcal{U}_i \in \mathbb{C}$ of receive antennas $i \in [1,2,3]$ of a three-channel radar, as
\begin{equation}
\mathcal{RD}( \textbf{p}_s ) =  10 \log_{10} \sum_{i=1}^3 \left|\mathcal{U}_i(\textbf{p}_s)\right|^2.
\label{eq:rdMap}
\end{equation}
For further information please refer to \cite{schroeder2013system}. We consider $\mathcal{RD}$ to refer to the totality of all pixels in the spectrum. \inblue{Alternatively, we could have chosen pointclouds from traditional radar perception instead as a guide for scene flow estimation. However, we selected the RD spectra, because (a) we wanted to be more independent from classical radar perception, and (b) radar pointclouds are obtained from RD spectra and therefore are subsets of those spectra. }
\subsection{Pixel sets for scene flow estimation}

Utilizing the instance segmentation information, the set of the camera's foreground pixels  is formed by those pixels $\mathbf{p}$ that have been identified to belong to one of the following object classes:
\begin{align}
	\mathcal{P}_{\text{fg}} \coloneqq\Big \{\mathbf{p} \Big | \mathcal{S}(\mathbf{p}) \in \{\text{pedestrian, car, truck, bicycle, motorbike}\}\Big\}
\end{align}
by the instance segmentation algorithm.
Further, let  $\mathcal{P}_\text{radar}$ be \inblue{the} set of camera pixels  covered by the radar's \gls{FOV}, which is defined as follows
\begin{equation}
\begin{split}
\mathcal{P}_\text{radar}\coloneqq \Big\{ \textbf{p}(\textbf{x}_R) \Big |  |\phi(\textbf{x}_R)|\leq \frac{135}{2}^\circ \wedge \theta(\textbf{x}_R)| \leq \frac{22}{2}^\circ\Big\}, \text{where}\\
\phi(\textbf{x}_R) = \tan^{-1}(x_{R,x},x_{R,y}), 
\theta(\textbf{x}_R) = \tan^{-1}(x_{R,x},x_{R,z}).
\end{split}
\end{equation}
Here, $\mathbf{x}_R$ is the point in the \gls{COOS} of the radar system, and  $\textbf{p}(\textbf{x}_R)$ the pixel in the camera image, corresponding $\textbf{x}_R$. 

To mitigate the effect of errors in instance segmentation, we perform further object clustering on each pointcloud belonging to an instance in camera image via DBSCAN \cite{Erst96}. Thereby, points which have a spatial distance larger than $\SI{0.3}{\meter}$ from the closest point in the cluster $i$ are classified as outliers and removed from the main cluster corresponding to the instance, resulting in the set $\mathcal{P}_{\text{DBSCAN}}$ of valid pixels. \inblue{The threshold of 0.3m  was considered appropriate for the used lidar sensor.}

For a pixel to be used for scene flow estimation, it must be both in the set of valid pixels after DBSCAN  clustering, $\mathcal{P}_{\text{DBSCAN}}$,  and in the \gls{FOV} of the radar sensor:
\begin{equation}
\begin{split}
\mathcal{P}_\text{i}\coloneqq \Big\{\textbf{p} \mid \textbf{p}\in
\mathcal{P}_{\text{DBSCAN}} \wedge  \textbf{p}\in \mathcal{P}_\text{radar}\Big\}.
\end{split}
\end{equation}
\inblue{
This set definition automatically handles the sensor parallax between radar and camera. An example of valid pixels is visualized in Fig. \ref{fig:pixelMasks_DoA_all_0} for both cameras. Valid pixels are highlighted as yellow pixels. It can be seen, that no pixel of the ego-vehicle was selected as valid. Similarly, only pixels from the radar FoV are marked as valid. Some pixels surrounding objects are automatically marked as invalid, according to $\mathcal{P}_{\text{DBSCAN}}$.
}

\inblue{

\begin{figure}[h]
  \centering
\includestandalone[width=0.5\textwidth]{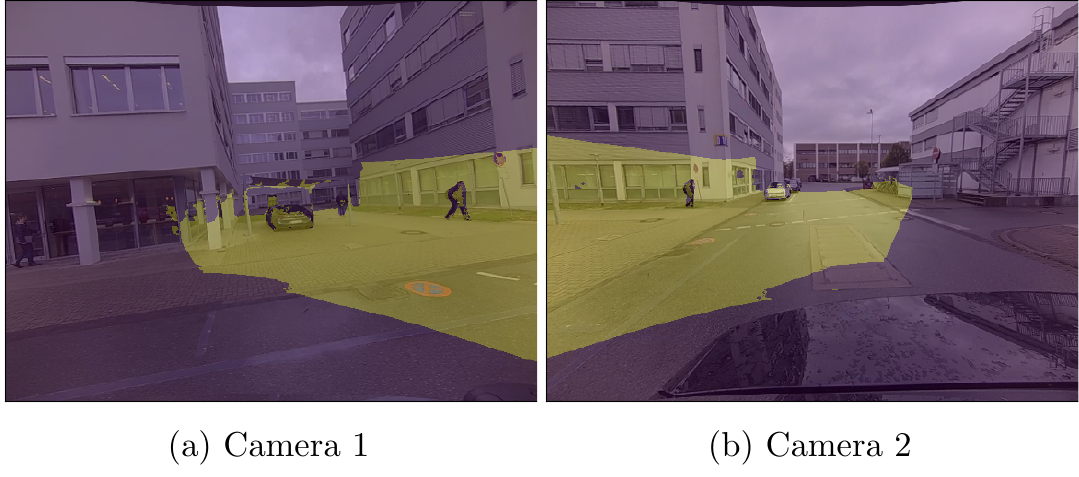}
\caption{\textbf{Masks for selection of valid pixels:} Examples of valid pixels (yellow) for all objects according to $\mathcal{P}_\text{i}$ visualized for both cameras.}
	\label{fig:pixelMasks_DoA_all_0}
\end{figure}

}

\subsection{Motion formulation}
The overall scene flow $\boldsymbol{\xi} = [\xi_x, \xi_y, \xi_z]^T$ is a superposition of background $\boldsymbol{\xi}_{bg}$ and foreground scene flow $\boldsymbol{\xi}_{fg}$,
\begin{equation}
\boldsymbol{\xi} = \boldsymbol{\xi}_{bg} + \boldsymbol{\xi}_{fg}.
\end{equation} 
The background scene flow encodes the motion of the ego-  $E$ over ground and applies to all pixels. Foreground motion is the motion of objects over ground. Similar to DRISF, the foreground motion is computed only for pixels in $\mathcal{P}_{\text{fg}}$. In contrast to DRISF, we decided to encode the foreground motion  in 3 translational motion parameters instead of 6 \mbox{(3 rotational + 3 translational)}. This simplification was made to improve the robustness and is justified by assuming small object dimensions and low rotational rates for foreground objects for the majority of traffic scenarios.

\subsubsection*{\textbf{Background motion formulation}}\label{ss:backgroundMotion}
The \inblue{perceived} motion of the background highly depends on the motion of the ego vehicle. Here,  rotation cannot be neglected since, even when the ego vehicle's rotation is small, the scene flow can be high for objects that are far away, see Eqs. \eqref{eq:pixelProjection} and \eqref{eq:sceneflow2} below.  

In the following, let $k$ denote the discrete time index.
Background motion is understood to be the relative 3-D motion of the environment, which is induced by the movement of the ego vehicle over ground. The position  $\textbf{x}_{E}{(\textbf{p}, k+1)}$ of a stationary point $\textbf{p}$ at time step $k+1$  can be predicted from its position $\textbf{x}_{E}{(\textbf{p}, k)}$ at time step $k$ as follows:
\begin{equation}
\textbf{x}_{E}{(\textbf{p}, k+1)} = \textbf{R}_{E, bg}\Big( \textbf{x}_{E}{(\textbf{p}, k)} + \textbf{x}_\text{RA}\Big) - \textbf{x}_\text{RA} - \textbf{t}_{E, bg},
\label{eq:pixelProjection}
\end{equation}
in which  $\textbf{R}_{E, bg}\in \rm I\!R^{3\times3}$ and  $\textbf{t}_{E, bg}\in \rm I\!R^{3\times1}$ describe the rotation and translation of the ego vehicle over ground, an information that can be obtained by \mbox{DGPS-INS}. In Eq.~\eqref{eq:pixelProjection}, before applying the rotation, the point $\textbf{x}_{E}{(\textbf{p}, k)}$ in the COOS of the ego vehicle is first  shifted onto the rear axle of the ego vehicle by adding $\textbf{x}_\text{RA}$, according to the Ackermann vehicle model \cite{Mitschke04}, which assumes that vehicles yaw around their rear axle. Similar as described in \cite{Menze2015CVPR}, $\textbf{R}_{E, bg}$ and $\textbf{t}_{E, bg}$ are refined over a sequence of camera images.

By   applying the coordinate transformation, Eq.~\eqref{eq:coos_transformation}, from ego to camera, we arrive at
\begin{IEEEeqnarray}{rCl}
\textbf{x}_C{(\textbf{p}, k+1)} &=& \textbf{R}_\text{C, bg} \textbf{x}_C{(\textbf{p}, k)} + \textbf{t}_\text{C, bg}\IEEEyesnumber\IEEEyessubnumber\label{eq:sceneflow2}\vspace{0.1cm}\\
\textbf{R}_\text{C, bg}(k) &=& {}^\text{C}\textbf{R}_\text{E} \textbf{R}_{E, bg} {}^\text{E}\textbf{R}_\text{C}  \IEEEyessubnumber\label{eq:sceneflow3}\vspace{0.1cm}\\
\textbf{t}_\text{C, bg} &=& {}^\text{C}\textbf{R}_\text{E} \textbf{R}_{E, bg} \textbf{x}_\text{RA} - 
{}^\text{C}\textbf{R}_\text{E} ( \textbf{x}_\text{RA} - \textbf{t}_{E, bg} )\nonumber\\&& + \>
\big(\textbf{I} -  \textbf{R}_\text{C, bg}(k)\big){}^\text{C}\textbf{t}_\text{E}.
%
%
%
%
%
%
\IEEEyessubnumber\label{eq:sceneflow4}
\end{IEEEeqnarray}
This eventually gives the background scene flow of point $\textbf{p}$:
\begin{equation}
\boldsymbol{\xi}_{bg}(\textbf{p}) = \textbf{x}_C{(\textbf{p}, k+1)} - \textbf{x}_C{(\textbf{p}, k)},
\label{eq:sceneflow}
\end{equation}


\subsubsection*{\textbf{Foreground motion formulation}}\label{ss:Energy formulation}

In \gls{DRISFwR}, a radar specific energy term is added to the energy formulation of DRISF:

\begin{small}
\begin{equation}
\begin{split}
\min_{\boldsymbol{\xi}} \Big\{
\underbrace{\lambda_{\text{photo}} E_{\text{photo}}(\boldsymbol{\xi}; \mathcal{I})+
\lambda_{\text{rigid}} E_{\text{rigid}}(\boldsymbol{\xi}; \mathcal{I})+
\lambda_{\text{flow}} E_{\text{flow}}(\boldsymbol{\xi}; \mathcal{I})}_{\text{original DRISF}}\\
+\underbrace{ \sum_{s_d =1}^3
\lambda_{\text{radar}, s_d} E_{\text{radar}}(\boldsymbol{\xi}; \mathcal{I}, s_d)}_{\text{extension by DRISFwR}} \Big\}.
\end{split}
\label{eq:energyFormulation}
\end{equation}
\end{small}
For the sake of simplicity, let \mbox{ $\mathcal{I} = \left\{\mathcal{RGB}^0, \mathcal{RGB}^1, \mathcal{D}^0, \mathcal{D}^1, \mathcal{S}^0, \mathcal{F}, \mathcal{RD} \right\}$}, thus any time an equation requires one or multiple of the arguments, we use $\mathcal{I}$ instead.
The $\lambda$'s are the weighting factors of the individual energy terms.  $E_{\text{photo}}$ describes, how well the $\mathcal{RGB}$ pixels from the object agree to those corresponding to $\mathcal{RGB}$, when predicted according to $\boldsymbol{\xi}$. $E_{\text{rigid}}$ describes how well pixel positions agree between frames assuming a scene flow of $\boldsymbol{\xi}$. The energy term $E_{\text{flow}}$ captures differences between the optical flow mask $\mathcal{F}$  and the obtained scene flow. For more details about these energy terms, we refer the reader to the original DRISF paper \cite{Ma19}, as we will focus on the radar specific extension here. 

The newly introduced $E_{\text{radar}}$ energy captures how well $\boldsymbol{\xi}$ agrees to the relative velocity measured by radar. 
This term is formulated as 
\begin{IEEEeqnarray}{rCl}
E_{\text{radar}}(\boldsymbol{\xi}; \mathcal{I}, s_d) &=& \sum_{\textbf{p} \in P_i} \rho\left( r_\text{radar}(\boldsymbol{\xi}, \textbf{p}; \mathcal{I}, s_d) \right)\IEEEyesnumber\IEEEyessubnumber\vspace{0.1cm}\\
r_\text{radar}(\boldsymbol{\xi},\textbf{p}; \mathcal{I}, s_d) &=& \mathcal{RD}_{s_d}\big(|\textbf{x}_R(\textbf{p})|, v_R(\boldsymbol{\xi},\textbf{p})\big) \nonumber\\ & & \qquad- \mathcal{RD}_\text{target},\IEEEyessubnumber
\label{eq:radarCosts}
\end{IEEEeqnarray}
where $\rho(r_\text{radar}) = \left( r_\text{radar}^2 + 10^{-6} \right)^{0.45}$ (generalized Charbonnier penalty) is the robust fitting function which we adopt from the original DRISF. The power of the RD-map value at range $|\textbf{x}_R|$ and relative velocity $v_R(\boldsymbol{\xi})$ seen from radar is $\mathcal{RD}\big(|\textbf{x}_R|, v_R(\boldsymbol{\xi})\big)$, whereas $\mathcal{RD}_\text{target}$ is the target value of the power and here set to be  the maximum power of the RD-map. This setting enforces the scene flow to maximize the RD-map power of an object in its local RD-map vicinity determined by the other energy terms. We found this behavior similar to what human labelers would expect. The summation index $s_d$, the scale level, will be discussed later in this section.

The minimization of \eqref{eq:energyFormulation} can be formulated as a weighted Least Squares (wLS) problem, see the supplementary material of \cite{Ma19} for details,
\begin{equation}
\boldsymbol{\xi}^{(m+1)} =  \underset{\boldsymbol{\xi}}{\mathrm{argmin}} \Big \{ \sum_{\textbf{p} \in \mathcal{P}_\text{i}} \textbf{r}^T(\textbf{p}, \boldsymbol{\xi}^{(m)}) \textbf{W}(\textbf{p}, \boldsymbol{\xi}^{(m)}) \textbf{r}(\textbf{p}, \boldsymbol{\xi}^{(m)}) \Big \}.
\label{eq:minStepFormulation}
\end{equation}
Here, $\textbf{r} \in \mathbb{R}^{K}$ is the vector of residuals, in which $K$ is the number of captured residual\inblue{s} from Eq.~\ref{eq:energyFormulation}. $\textbf{W} \in \mathbb{R}^{K \times K}$ is the diagonal weight matrix  which is specified by the $\lambda's$ and the selected fitting function as shown later in this section.

%

In each iteration $m$, the \gls{GN} solver performs scene flow updates \mbox{$\boldsymbol{\xi}^{(m+1)} = \boldsymbol{\xi}^{(m)} + \Delta \boldsymbol{\xi}^{(m)}$} by computing 
\begin{equation}
\Delta \boldsymbol{\xi}^{(m)} = -\left( \sum_{\textbf{p} \in \mathcal{P}_\text{i}} \textbf{J}^T \textbf{W} \textbf{J} \right)^{-1} \sum_{\textbf{p} \in \mathcal{P}_\text{i}} \textbf{J}^T \textbf{W} \textbf{r}.
\label{eq:gnStep}
\end{equation}
Here, $\textbf{J} \in \mathbb{R}^{K \times 3}$ is the Jacobian giving the  gradient of the residuals in each scene flow direction. 
All terms on the right hand side depend on $\textbf{p}$ and $\boldsymbol{\xi}^{(m)}$, which we omitted, however,  for the sake of readability.

The Jacobian $\textbf{J}_\text{radar} \in \mathbb{R}^{1x3}$ corresponding to the residual of the radar term $r_\text{radar}$, which defines one row in $\textbf{J}$, is computed as follows:
\begin{equation}
\textbf{J}_\text{radar} = \frac{\partial r_\text{radar}(\boldsymbol{\xi}; \mathcal{I}, s_d)}{\partial \boldsymbol{\xi}} = 
\frac{\partial \mathcal{RD}(\boldsymbol{\xi}; \mathcal{I}, s_d)}{\partial v_r(\boldsymbol{\xi})}
\frac{\partial v_r(\boldsymbol{\xi})}{\partial \boldsymbol{\xi}_\text{radar}(\boldsymbol{\xi})}
\frac{\partial \boldsymbol{\xi}_\text{radar}(\boldsymbol{\xi})}{\partial \boldsymbol{\xi}}.
\label{eq:Jacobian}
\end{equation} 

\begin{figure}[h]
  \centering
\includestandalone[width=0.41\textwidth]{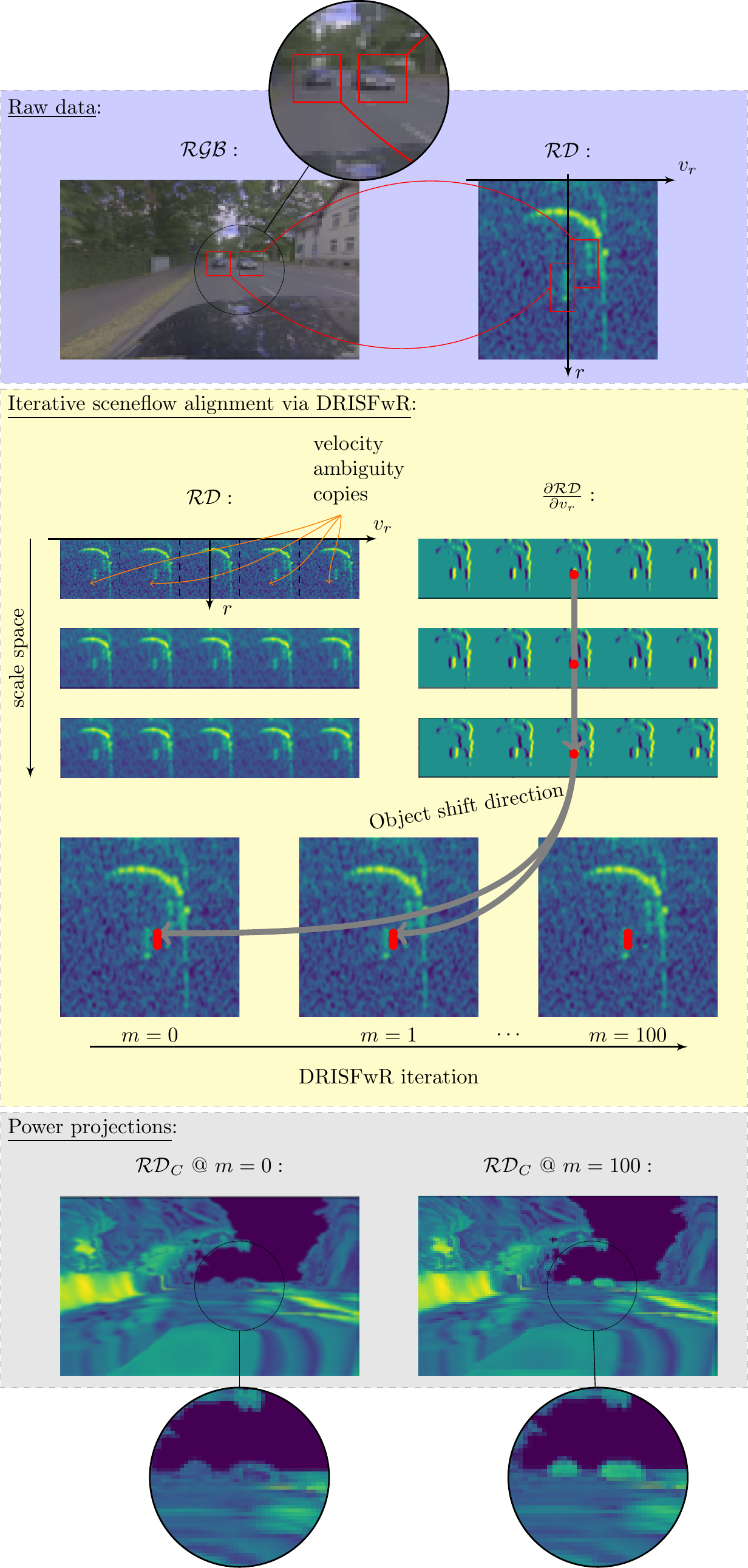}
 \caption{\textbf{Automatic scene flow alignment to Radar data via \mbox{DRISFwR}:} Top: RGB image and RD-map with two vehicles. Middle: Scale-space of radar data used in DRISFwR with energy (left) and partial derivative (right). One object is visualized in scalespace as red point. Repositioning via scene flow adjustment in consecutive DRISFwR update steps with gray arrows marking shift direction. Bottom: Power projections. After DRISFwR convergence, energy projection from RD-map is captured correctly for both vehicles (bottom right)}
    \label{fig:drisfwrProcess}
\end{figure}

The first term on the right hand side represents the dependency of RD-map power on relative radial velocity in form of image gradients. Example gradients are illustrated in Fig. \ref{fig:drisfwrProcess}, the right-hand figures in the box with the yellow background.  At RD-map positions with high power, one can see large negative or positive gradients, the sign depending on the horizontal approaching direction.

Since the image gradients in the RD-map are very steep, a \mbox{scale-space} \cite{Lin94} is created by repeated Gaussian smoothing, max-pooling of $\mathcal{RD}$ and bilinear up-sampling along the Doppler dimension. The smoothed RD-map is referred to as $\mathcal{RD}_{s_d}$. A Gaussian kernel is applied to obtain smooth gradients. Max-pooling enhances the receptive field of DRISFwR, which allows object alignment in RD-map over a wider range of velocities. Upscaling is performed to keep the bin resolution constant over scales and to induce further content smoothing. Different scale levels $s_d$ are considered in DRISFwR by accumulating their energy in Eq. \eqref{eq:energyFormulation}. To prefer alignment on lower scales, energy weighting $\lambda_\text{radar}$, is bisected at every scale level $s_d$.  Furthermore, to cope with aliasing in the Doppler spectrum, copies of the radar spectrum are concatenated in Doppler direction, see \mbox{Fig. \ref{fig:drisfwrProcess}}.

The second term on the right hand side of Eq. \ref{eq:Jacobian} incorporated the scalar projection of the scene flow onto positional vector $\textbf{x}_R$, both observed from radar perspective.
\begin{equation}
v_r(\boldsymbol{\xi}_\text{radar}) = \frac{\textbf{x}_R^T}{|\textbf{x}_R|} \boldsymbol{\xi}_\text{radar}, 
\label{eq:relativeVelocity}
\end{equation} 
Its derivative is given by:
\begin{equation}
\frac{\partial v_r(\boldsymbol{\xi}_\text{radar})}{\partial \boldsymbol{\xi}_\text{radar}} = \frac{\textbf{x}_R^T}{|\textbf{x}_R|}.
\end{equation} 
The relation between a position in the radar COOS and in the camera COOS is given by coordinate transformation according to 
Eq.~\eqref{eq:coos_transformation}. Thus
\begin{IEEEeqnarray}{rCl}
\boldsymbol{\xi}_\text{radar} &=& \frac{\partial \textbf{x}_R}{\partial t} = {}^R\textbf{R}_E {}^E\textbf{R}_L {}^C\textbf{R}_L^{-1} \frac{\partial \textbf{x}_C}{\partial t} \nonumber\\ &=& {}^R\textbf{R}_E {}^E\textbf{R}_L {}^C\textbf{R}_L^{-1} \boldsymbol{\xi} = {}^R\textbf{R}_C \boldsymbol{\xi},
\label{eq:sceneflowInRadarCoos}
\end{IEEEeqnarray}
which transforms the scene flow from camera coordinates into radar coordinates via the rotational matrix ${}^R\textbf{R}_C$. We obtain 
\begin{equation}
\frac{\partial \boldsymbol{\xi}_\text{radar}}{\partial \boldsymbol{\xi}} = {}^R\textbf{R}_C.
\end{equation} 

The adaptive weighting ${W}_\text{radar}$ is given by:
\begin{equation}
{W}_\text{radar} = \lambda_{\text{radar}}  \frac{\partial^2 \rho}{\partial {r}_\text{radar}^2} = 0.45 \lambda_{\text{radar}}\left( {r}_\text{radar}^2  + 10^{-6} \right)^{-0.55}.
\label{eq:confidence}
\end{equation}

An update step of the \gls{GN} algorithm is depicted in Fig. \ref{fig:drisfwrProcess}. The figure shows two cars highlighted by the red rectangles in the camera image and the RD-map in the blue box. In the yellow box, the position in the RD-map for one car is shown in red for multiple update steps. After 100 iterations, the car's position in RD-map has shifted into the local power maxima in the RD-map. The remaining gray box shows the RD-map power warped into the camera image. At the location of the cars in the camera image, the warped power has significantly increased after 100 iterations. The warping operation will be discussed in the next section.

GN iterations are stopped  once $\|\Delta \boldsymbol{\xi}\|_2< \SI{0.1}{\milli\meter}$. Typically this occurred after approx. 30 iterations but was stopped when convergence was not reached after 100 iterations.

\section{Warping from radar spectra into camera image}\label{ss:warping}
In the previous section, we described the  scene flow  and the dense depth estimation. As can be seen in Fig. \ref{fig:ablaufdiagram3}, they form the fundamental input for the warping operation which we will describe in the following.

\subsection{Bilinear warping operation}\label{ss: bilinearWarping}
\inblue{The radar RD grid is given by coordinates of radial relative velocity $v_r$ and range $|\textbf{x}_r|$ seen from the radar perspective. These can be obtained following  Eq.~\eqref{eq:coos_transformation} and Eq.~\eqref{eq:relativeVelocity}. Note, that only the radial relative velocity is relevant for the warping operation, as the radar sensor is not able to measure tangential velocities.}

 In our warping operation $\eta(\cdot,\cdot,\cdot)$,  we select for each pixel in the camera image the corresponding pixel in the RD grid depending on the assigned $v_r(\boldsymbol{\xi})$ and $|\textbf{x}_r|$ and project the assigned RD grid pixel value back into the camera image grid. If $v_r(\boldsymbol{\xi})$ and $|\textbf{x}_r|$ are not on the RD grid,  bi-linear interpolation is applied, since it provides visually appealing projections while being differentiable, which enables the backpropagation in  \gls{NN} training. 

\begin{figure}[h]
  \centering
\includestandalone[width=0.5\textwidth]{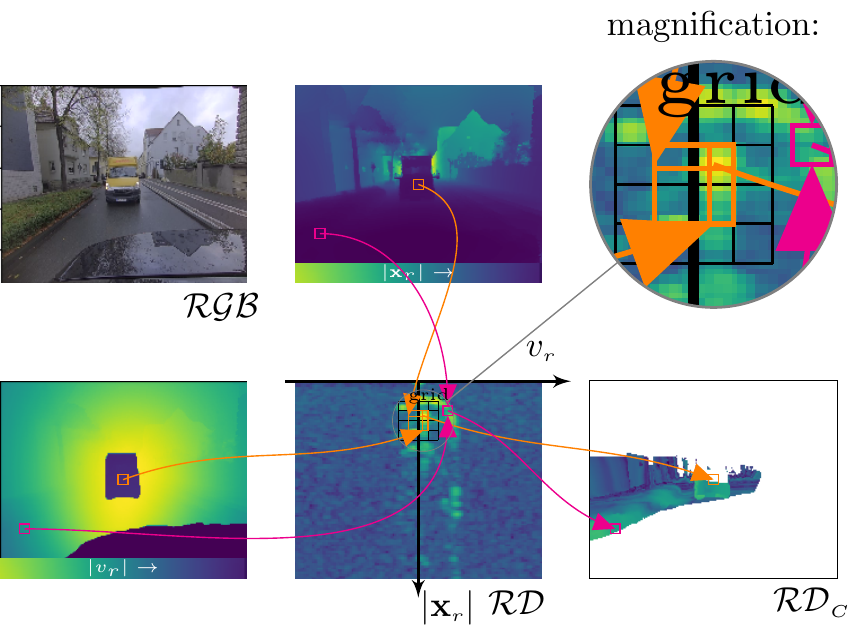}
  \caption{\textbf{RD-map warping into camera image:}     As an example for the warping operation, here for two pixels in the camera image (upper left) their estimated range (upper center) and radial velocity (lower left) determines their corresponding position in the RD-map (lower center). The RD-maps intensity value is bi-linearly interpolated (upper right) and then projected into the camera image (lower right).
    Notice that only pixels in  the radar's FoV are warped here (non-white pixels in $\mathcal{RD}_C$).}
    \label{fig:bilinearInterpolation}
\end{figure}

Fig. \ref{fig:bilinearInterpolation} shows an example of  warping  an RD-map into the camera image. Such a warping helps a human to understand how a radar sees the environment in terms of power. We will refer to the projected RD-map as $\mathcal{RD}_C$ in the following. 

$\mathcal{RD}_C$ can be calculated at the pixel position \inblue{$\textbf{p}$}, using bilinear interpolation, as:
\begin{equation}
\mathcal{RD}_C(\textbf{p}) = \eta\Big( \mathcal{RD}, v_r(\boldsymbol{\xi}), |\textbf{x}_r|, \textbf{p} \Big).
\label{eq:warpingEq2}
\end{equation}

Note, that in Eq. \eqref{eq:warpingEq2} and Fig. \ref{fig:bilinearInterpolation} the RD-map $\mathcal{RD}$ was depicted as an illustrative example for the warping operation. However, every information in the form of the RD grid can be warped into a camera image in an analog fashion. In fact, we will warp a radar based prediction by NNs in the same manner, as will be discussed in the remainder of this paper. 

\inblue{
\subsection{Bilinear vs. trilinear warping}\label{ss: bilinearVsTrilinearWarping}
In the bilinear interpolation only velocity and distance from our reference are used for warping. But the proposed framework is flexible and can also employ 3-D data for warping. To demonstrate that,  Fig. \ref{fig:warping_comparison_all_0} shows the warping of a 3-D radar spectrum, which consists of velocity, distance and azimuth DoA information, into a camera image. The 3-D radar spectrum was obtained by performing FFT beamforming \cite{Krim96} on the radars channel returns. For warping, trilinear interpolation was used if the considered camera pixel does not correspond to a point on the 3-D RD grid.


\setlength{\figW}{0.45\columnwidth}
\setlength{\figH}{0.75\figW}

\begin{figure}[h]
  \centering
\includestandalone[width=0.48\textwidth]{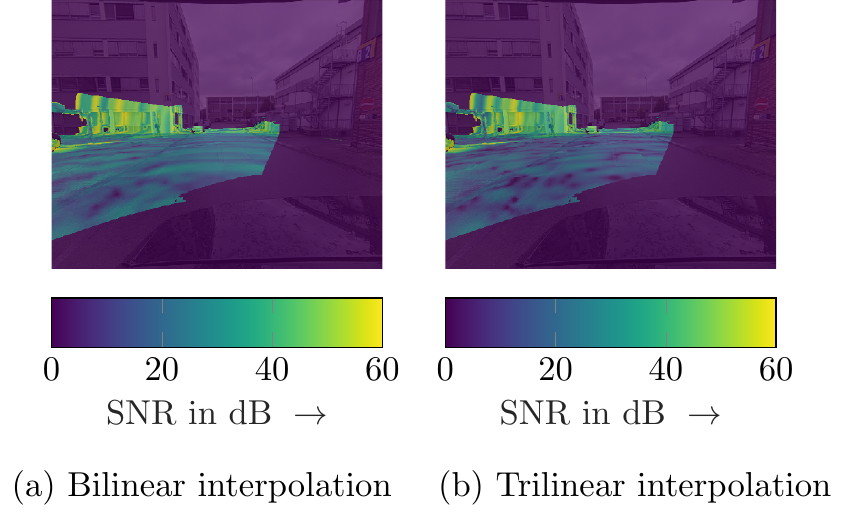}
\caption{\textbf{Comparison of the interpolation methods:} Left: Projection of the power from RD-map via bilinear interpolation. Right: Projection of the 3D beamforming spectrum via trilinear interpolation. Pixels outside the radars FoV are visualized dark.}
	\label{fig:warping_comparison_all_0}
\end{figure}

We normalized the signal power in the beamfroming spectrum according to the number of used antennas, thus highly reflective regions in Fig. \ref{fig:warping_comparison_all_0} do have similar power values. The main difference in the two images can be found in the low power regions, e.g. the street area, in which better noise suppression was achieved by beamforming. 
}

\section{Cross-modal supervised NN training on radar spectrum data}\label{ss:appliedSupervision}

In this section, we describe an example application for training a ``student'' neural network operating on radar spectrum data, whose training targets have been provided from a ``teacher'' neural network, which labeled the camera data.



\subsection{Radar based DoA prediction via Neural Network}

To estimate the DoA of the radar reflections, a \gls{CNN}  is used to operate on the radar spectrum ~$\mathcal{U}$. To be precise, the radar spectrum of the utilized three-channel radar $\{\mathcal{U}_1, \mathcal{U}_2, \mathcal{U}_3\}$  is preprocessed before being transferred to the network. The first feature map is the RD-map, while the second and third feature maps are the phase spectra $\angle \left( \mathcal{U}_i, \mathcal{U}_{i-1} \right)$ between adjacent channels:

 \begin{IEEEeqnarray}{rCl}&&
\Gamma =
\begin{bmatrix} 
 \mathcal{RD}\\ \angle \left( \mathcal{U}_2, \mathcal{U}_{1} \right)
\\ \angle \left( \mathcal{U}_3, \mathcal{U}_{2} \right) ,
\end{bmatrix}\IEEEyesnumber\IEEEyessubnumber\vspace{0.1cm}\\
&& \angle \left( \mathcal{U}_i, \mathcal{U}_{i-1} \right)  = \arg(\mathcal{U}_i \mathcal{U}{^\ast}_{i-1}).\IEEEyessubnumber
\label{eq:warpingEq}
\end{IEEEeqnarray}
\inblue{
The RD-map is provided to allow the neural network to discriminate between targets and noise and the phase spectra are provided as a main clue for DoA inference.
}

For every pixel in the RD grid, the NN, denoted as $\phi\text{-Net}(\Gamma)$, delivers a DoA estimate $\phi_\text{RD-grid}$ in azimuth direction which is warped into the camera image as $\phi_\text{pred., RD}$:
\begin{IEEEeqnarray}{rCl}
\mathcal{\phi}_\text{pred., RD} &=& \phi\text{-Net}(\Gamma)\\
\phi_\text{pred., cam}(\textbf{p}) &=& \eta\Big(\phi_\text{pred., RD} , v_r(\boldsymbol{\xi}), |\textbf{x}_r|, \textbf{p}\Big).
\label{eq:doaPred}
\end{IEEEeqnarray}
For the network training, DoA labels $\phi_\text{reference}$ are generated from the dense depth estimation in the camera image, projected into the radar \gls{COOS} as: \begin{equation}
  \phi_\text{reference} = \arctan2(x_{R,x}, x_{R,y} ).
\end{equation}
Predictions of $\phi$-Net are made, so that the objective $l_{\phi-\text{Net}}$ is minimized during network training:

\begin{equation}
\label{eq:lossFunction_DoA_0}
l_{\phi-\text{Net}} = \sum_{\textbf{p} \in \mathcal{P}_\text{train} }
\big|
\phi_\text{reference}\left(\textbf{p}\right) - \phi_\text{predict, cam}\left(\textbf{p}\right)
\big|, 
\end{equation}
  in which the summation is done over all pixels $\textbf{p}$ in $\mathcal{P}_\text{train}$, as we want the network to focus on relevant targets. Examples for $\phi_\text{reference}$, $\phi_\text{predict, cam}$ and $\phi_\text{predict, RD}$ are depicted in Fig. \ref{fig:azimuthAngle2}. Note, that $\phi_\text{reference}\left(\textbf{p}\right)$ and $\phi_\text{predict, cam}\left(\textbf{p}\right)$ provide angles within sensor FoV $<180^\circ$ by physical properties and therefore $2\pi$ wrap must not be treated.

\section{Evaluation}\label{ss:nnEvaluation}

\subsection{Data Acquisition System}\label{ss:measurement_setup}

\begin{figure}[h]
  \centering
\includestandalone[width=\columnwidth]{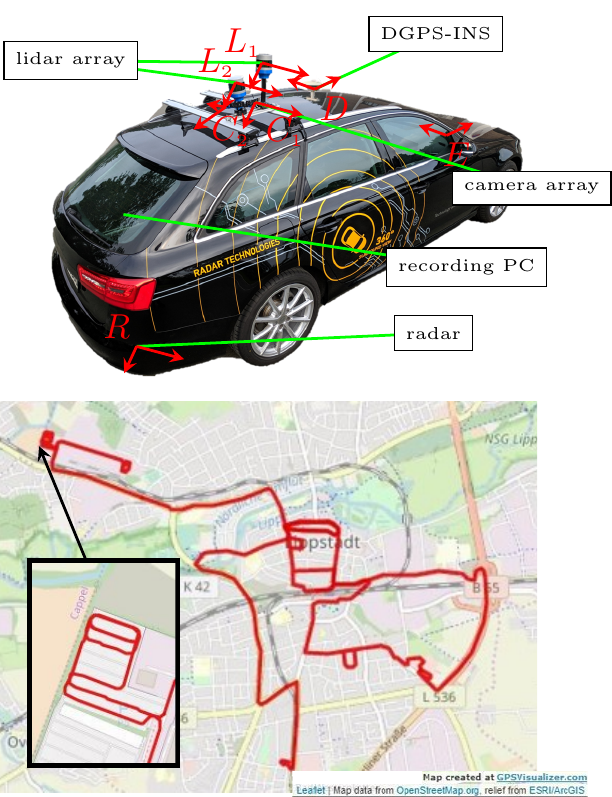}
   \caption{\textbf{Data capturing setup:} Top: vehicle sensor configuration; bottom: driving trajectory for recording the database; bottom left: magnified scene flow evaluation area.}
    \label{fig:vehicleConfiguration}
\end{figure}

To gather real world sensor data from driving scenarios, we equipped a car with a 77 GHz HELLA radar sensor, with an overview of the sensor specifications given in Tab. \ref{tab:radarSpecs}. The sensor captures the raw data, computes RD spectra in the first processing step and forwards them to an in-vehicle recording PC.  The radar sensor is mounted at the rear right corner  and behind the bumper, as illustrated in  Fig. \ref{fig:vehicleConfiguration}. 

\inblue{
\begin{table}[h]
\centering
\caption{\textbf{Specifications of the tested radar sensor}}
\label{tab:radarSpecs}
\begin{tabular*}{0.45\textwidth}{ll}
\toprule
Operating frequency & 77 GHz \\
Antenna design & static 1Tx3Rx\\
Modulation scheme & FMCW chirp sequence\\
Range resolution & \SI{0.25}{\meter}\\
Doppler resolution & \SI{0.25}{\meter\per\second}\\
Maximum range & \SI{25}{\meter}\\
Maximum unambiguous Doppler & \SI{10}{\meter\per\second}\\
FoV (horizontal/vertical) & $140^\circ/20^\circ$\\
\bottomrule
\end{tabular*}
\vspace{0.1cm}
\end{table}
}

The utilized reference sensor system consisting of 2 Velodyne VLP-32C lidar scanners \cite{velo20} at 10 FPS and a camera array (2 \mbox{FirstSensor DC3C-1-E4P-105} \cite{FS20}) is mounted on a luggage rack on top of the vehicle. In order to sample the surrounding more densly, the system is equipped with two lidar sensors, which are mounted with different orientations on the vehicle.  They jointly provide a denser point-cloud in the radar's FoV than a single lidar could provide, see Fig. \ref{fig: depthCompletion}. In order to cover the entire radar FoV (azimuth: $135^\circ$, elevation: $22^\circ$), two cameras (azimuth: $\approx 104^\circ$) with overlapping FoVs were used. The camera array is able to record a continuous videostream at 30 FPS. The Differential-GPS with inertial navigation system (DGPS-INS) of type GeneSys ADMA-G-Pro+ \cite{Gen20} is used as a reference sensor for precise vehicle over ground motion estimation. \inblue{
All sensors have been properly synchronised via temporal calibration proposed in \cite{Merriaux17}.} \ingreen{To be specific, the trigger times of the sensors have been recorded and the temporally closest sensor samples were associated. Additionally, motion correction was applied on the data of the  lidar sensors. }

The driving trajectory for this dataset is given in \mbox{Fig. \ref{fig:vehicleConfiguration}} and contains mostly urban environment. It has been recorded over a period of approx. $\SI{1}{\hour}$ driving, giving rise to 36000 frames at 10 FPS and approx. 7 billion samples on pixel level.

\subsection{Scene flow evaluation}\label{se:sceneflowEval}
The performance of the proposed scene flow algorithm has been benchmarked on a subset of our dataset, with known ground truth for scene flow. A parking lot area (see Fig. \ref{fig:vehicleConfiguration}) with all objects being stationary, was selected. The ground-truth for scene flow estimation was obtained from DPGS-INS, which delivered precise information about the ego motion of the vehicle.  
Scene flow in stationary scenes has been described in Sec. \ref{ss:backgroundMotion}, and thus the scene flow ground truth $\boldsymbol{\xi}_{gt}$ is  given by by $\boldsymbol{\xi}_{gt} = \boldsymbol{\xi}_{bg}$.

The \textit{KITTI} dataset provides scene flow evaluation metrics mainly in camera image plane direction. In our scenario, scene flow errors both orthogonal and tangential to the image plane are important and thus we decided to use error metrics that consider both errors equally. The first is the \gls{MAE}
\begin{equation}
\begin{split}
\text{MAE}_{sf} = \frac{1}{N} \sum_{\textbf{p}_{sf} \in P_{sf}} \|\boldsymbol{\xi}_{gt}(\textbf{p}_{sf}) - \boldsymbol{\xi}(\textbf{p}_{sf})\|_2^1 \\ 
\mathcal{P}_{sf} \coloneqq \{ \textbf{p} \mid \textbf{p}\in P_\text{radar} \wedge  \textbf{p} \in P_\text{sparse} \wedge \textbf{p} \in P_\text{fg}\},
\end{split}
\end{equation}
between ground truth and estimated scene flow, measured on all pixels $\textbf{p}$ visible to the radar, have assigned lidar sparse depth measurement and have an instance mask (i.e., are foreground pixels). Here, $N$ is the cardinality of that set $\mathcal{P}_{sf}$.
 
The second evaluation metric is the scene flow error rate, which is the number of foreground pixels exceeding an error threshold of $\SI{0.25}{\meter/\second}$, which is a typical radar velocity resolution.

As can be seen in Tab. \ref{tab:sceneflowCompare} the DRISFwR algorithm achieves lower \gls{MAE} and error rates than the current top-ranking scene flow estimation algorithm in \textit{KITTI} benchmarks. Both algorithms assign a strong motion prior, by assuming rigid bodies for each instance from instance segmentation. As an alternative scene flow estimation technique with weak motion constraints, we include an estimation based on optical flow by warping depth images between frames via $\text{HD}^{3}$ \cite{Yin19}. It can be seen, that scene flow performance without strong motion prior assumption is much worse. This is expected, as the nominal accuracy of the reference system is $0.16^\circ$ tangential and $\SI{0.03}{\meter}$ radial giving a theoretical three-dimensional accuracy of $\approx \SI{1.42}{\meter\per\second}$ for a target in $\SI{20}{\meter}$ distance. The rigid object assumption often results in hundreds of pixel level samples per object allowing to stay below the nominal accuracy of the reference sensors itself. The runtime of DRISFwR exceeds the one of DRISF slightly, as it includes an additional energy term. The computation is performed offline, allowing us to automatically label our $\SI{1}{\hour}$ long dataset within 1 day, which we found acceptable. \ingreen{All algorithms have been implemented in PyTorch \cite{NEURIPS2019_9015}, and the GPU code has been executed on NVIDIA GeForce GTX TITAN X.}

\begin{table}[h]
\centering
\caption{\textbf{Quantitative comparison of scene flow approaches}. DRISFwR reduces scene flow MAE and error rate compared to DRISF and the an estimator without strong motion prior $\text{HD}^{3}$.}
\label{tab:sceneflowCompare}
\begin{tabular*}{0.45\textwidth}{lccc}
\toprule
Methods & runtime & error rate (\%) & $\text{MAE}_{sf}$ ($\SI{}{\meter/\second}$)\\
\midrule
$\text{HD}^{3}$ & 0.12 sec & 69.9 & 4.37 \\
DRISF & 0.6 sec & 31.2 & 0.31 \\
DRISFwR & 0.8 sec & 25.5 & 0.22\\
\bottomrule
\end{tabular*}
\vspace{0.1cm}
\end{table}

 Note, that for our application the radial velocity error is likely to be smaller than the  values given in the table, as we only require accurate radial velocity estimates, whereas the metric measures both radial and tangential velocity deviations.


\subsection{Automatic labeling quality}\label{ss:labelEval}
\begin{figure}[h]
  \centering
\includestandalone[width=0.48\textwidth]{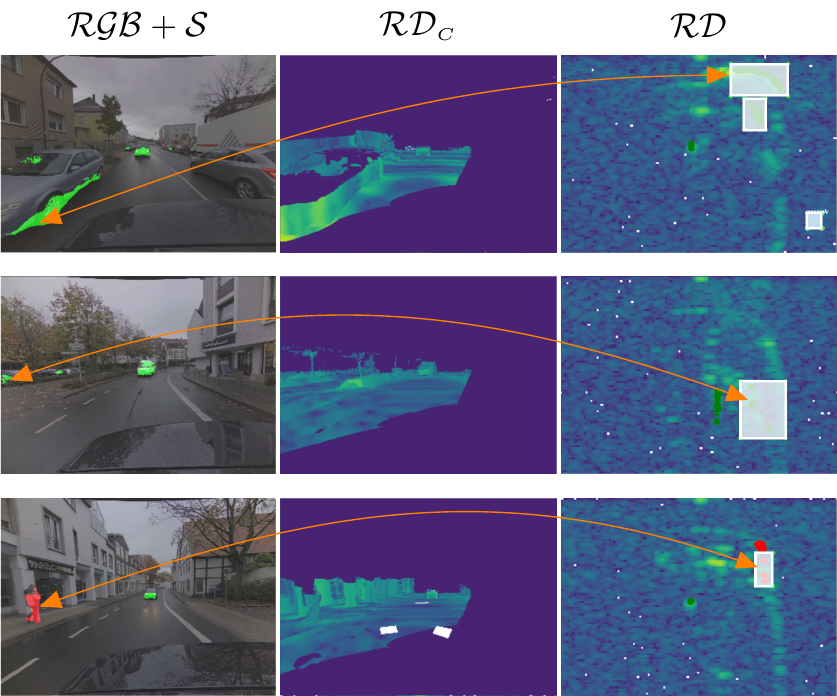}
    \caption{\textbf{Examples of anomalous labels:} From left to right: camera images with semantic instance masks from radar FoV only (red: pedestrian, green: vehicle), projected RD-maps, RD-maps with semantic masks and anomaly masks (white rectangles). The orange arrows connect corresponding regions in camera image and RD-map. In the two upper rows, stationary objects were labeled as moving, due to an error in scene flow estimation (error type 1). However, the misalignment in RD-map is relatively small. The bottom row shows an error in instance segmentation, in which background pixels were mapped as pedestrian, resulting in an unreasonably extended object range in RD-map (error type 3).}
    \label{fig:anomalyMasks}
\end{figure}

DRISFwR assists the warping of radar spectra into camera images without human supervision, within the scope of the evaluated accuracy, see prior section. However, sometimes scene flow is found to be erroneous. Three error types can be discerned: 
\begin{enumerate}
\item stationary objects are mis-labeled as moving
\item significant portions of the RD-map power is not transferred into $\mathcal{RD}_C$
\item erroneous segmentation of point cloud resulting in unreasonable geometric extension in $\mathcal{RD}$. 
\end{enumerate}
Fortunately, these types of errors can easily be spotted by a trained human and labeled as anomalies and thus be excluded from the subsequent comparison between the reference data and the estimates from radar. Thus, with the help of our reference label generation system, the human labeling task can be tremendously simplified from a regression problem (draw DoA reference angles) to a binary classification problem (identify whether proposed label is plausible or not). We think, the first problem cannot even be done by human labelers, while the later one can easily be done. Our test dataset (10800 images) has been labeled by following this process. Examples of anomalous masks are given in \mbox{Fig. \ref{fig:anomalyMasks}}. Of all test dataset images, approx. $20.3\%$ received at least one anomaly mask. We consider this value reasonable, as it is close to the warping error rate for DRISFwR, identified in the last section. The average human labeling time per image was approx. $\SI{5}{\second}$. As the anomaly labeling catches all anomalies induced by erroneous warping, the anomaly masks can be used not only for the DoA application, but also for  training other applications, such as target detection and semantic segmentation. 

Note that the human anomaly labeling has only been applied to the test dataset in order to produce the most reasonable metrics. The \gls{NN} training, however, was performed without this human intervention and thus includes even warping errors. 

\subsection{Direction-of-arrival evaluation}\label{ss:doaEval}

Typically, the radar based estimation quality highly depends on the signal-to-noise-ratio (SNR). We therefore measured the performance as a function of the SNR. Here, the signal power is estimated as the power of the corresponding pixel in $\mathcal{RD}$. The noise level estimate is obtained 
by applying the CFAR target detector \cite{schroeder2013system} to $\mathcal{RD}$ and calculating the $99.5\%$ percentile of all non-target pixels.
In stationary scenes with low ego vehicle velocity $<\SI{2}{\meter\per\second}$, the signal power from multiple reflections accumulates in the zero Doppler velocity pixels. These signal powers cannot be recovered, leading to erroneous signal power estimation and therefore, deteriorate the quality metrics unreasonable. Stationary scenes are thus discarded from the test dataset.

The warping operation is non-bijective in the sense that multiple pixels in the camera may correspond to a single pixel in the RD-map and vice versa. In evaluation, this typically means (as the former case is more typical), that one estimation value (one pixel in the RD grid) is compared to an ensemble of reference values $\mathcal{P}_s$ (multiple pixels in the camera image). The set $\mathcal{P}_s$ can be formed by warping the linear pixel indices of RD-map into camera image and selecting all pixels to a certain pixel indice value. Ideally, one would want to assign an electromagnetic contribution to each reference pixel and compute the center of reflection to which the estimated value is then compared to. However, at the current stage of this reference system, the electromagnetic contribution is unknown. In practice, we select the best matching pixel from the camera ensemble to the RD grid pixel as the tuple $\big\{\textbf{p}_C, \textbf{p}_\text{RD}\big\}$ such that
\begin{equation}
\textbf{p}_{C} = 
\argmin_{\textbf{p}_C \in \mathcal{P}_s}
\Big|\phi_\text{reference}\left(\textbf{p}_C\right) - \eta\Big(\phi_\text{predict, RD}(\textbf{p}_{RD}); v_r, |\textbf{x}_r|\Big)\Big|.
\end{equation}
The DoA \gls{MAE}  is thus computed as 
\begin{equation}
\text{MAE}_\text{DoA} = \frac{1}{N}
\sum_{\textbf{p}_C \in \mathcal{P}_\text{radar}}
\Big|\phi_\text{reference}\left(\textbf{p}_C\right) -  \eta\big(\phi_\text{predict, RD}(\textbf{p}_{RD}); v_r, |\textbf{x}_r|\big)\Big|,
\end{equation}
where $N$ is the number of RD grid pixels visible to camera and radar, respectively the size of the set $\mathcal{P}_\text{radar}$.

We developed a \gls{CNN} for DoA estimation. It employs ReLU activation on every layer except for the last, where $90\cdot\tanh()$ is applied to scale the logits angular values to degrees. The convolutions are operated with unit stride and the layers have the following number of output channels: $[3, 32t, 64t, 128t, 64t, 32t, 32t, 1]$. \inblue{We experimented with different convolution kernel sizes of $1\times1$ and $3\times3$ and equalized the networks parameter count with the layer modifier $t$. The inference of the $1\times1$ network will be influenced by one pixel in the RD grid only, while the $3\times3$ network may use the RD grid pixel vicinity to perform inference. The receptive fields thus compute to 1 ($\SI{0.25}{\meter}\times\SI{0.25}{\meter\per\second}$) and 15 ($\SI{3.75}{\meter}\times\SI{3.75}{\meter\per\second}$), respectively. }

\inblue{
We subdivided the pixels in RD map into two groups, the first being pixels from real reflections, thus carrying a noisy signal, and the second  being pixels from noise only. As we want to focus the network training on the first group only, we automatically identify pixels of the second group as those whose SNR is found to be below the threshold of $10\text{dB}$, giving rise to the following definition of the training set
\begin{equation}
\mathcal{P}_\text{train} \coloneqq \{ 
\textbf{p} \mid \textbf{p}\in P_\text{radar}
\wedge  
\mathcal{RD}_C(\textbf{p}) > 10\text{dB}
\}.
\end{equation}
This threshold has been found by plotting the distribution of $\mathcal{RD}_C$ for the entire test dataset, see Fig. \ref{fig:snrDistribution_0}, and identifying the SNR value in which the likelihood of the second group samples can be neglected. To this end, a two-component mixture model was fitted to the data, consisting of a Gaussian and a Chi-squared distribution. The parameters of the Gaussian \gls{PDF}, that represents noise, was determined according to central limit theorem \cite{durrett_2019}. Assuming \mbox{Swerling type 3} fluctuations for the radar reflections, see e.g. \cite{Mes06}, the parameters of the Chi-squared \gls{PDF} representing the signal were identified.


As can be seen, at  $+10\text{dB}$ the likelihood of sampling from the identified noise mixture is very low ($<0.1\%$) and samples above $+10\text{dB}$ are mainly provided by signal.


\begin{figure}[h]
 \centering
\begingroup
 \inputencoding{latin1} 
\includestandalone[width=0.48\textwidth]{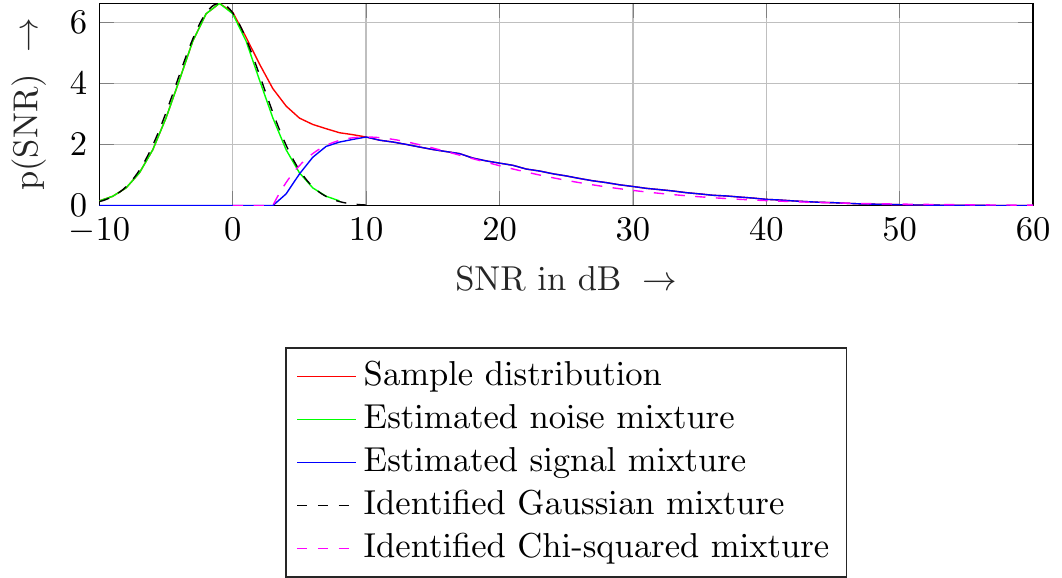}
\endgroup
    \caption{\textbf{SNR distribution:} 
The sample distribution is shown in red, which is decomposed into a noise component (green) and a signal component (blue). Additionally, the sample mixtures are approximated by Gaussian \gls{PDF} (dashed black) and Chi-squared \gls{PDF} (dashed magenta).}
\label{fig:snrDistribution_0}
\end{figure}

To explore the influence of the above mentioned threshold selection, we tested each network either with or without the SNR-based sample selection. The parameters of the three tested \gls{NN} architectures are depicted in Tab. \ref{tab:nnArchitectures}.
}

\begin{table}[h]
\centering
\caption{\textbf{NN architectures and their \inblue{configurations}}.}
\label{tab:nnArchitectures}
\begin{tabular*}{0.45\textwidth}{lccc}
\toprule
Name & kernel size & layer modifier (t) & \inblue{SNR selection} \\
\midrule
$\text{NN}_0$ & $1\times1$ & 3 & - \\
$\text{NN}_1$ & $1\times1$ & 3 & $\mathcal{RD}_C  > 10\text{dB}$\\ 
$\text{NN}_2$ & $3\times3$ & 1 & - \\
$\text{NN}_3$ & $3\times3$ & 1 & $\mathcal{RD}_C>10\text{dB}$ \\
\bottomrule
\end{tabular*}
\end{table}

As we want the training and test datasets to be independent and identically distributed (i.i.d.), the $\SI{1}{\hour}$ long dataset \mbox{(Fig. \ref{fig:vehicleConfiguration})} is split into \SI{10}{\second} long sequences of 100 frames each. The sequences are randomly assigned to training, validation and test sets with ratios $70\%, 15\%, 15\%$ and with no frame overlap.


All networks are optimized via ADAM and a learning rate of $10^{-5}$. Early stopping was performed by checking network performance on the validation dataset after each training epoch.

\begin{figure}[h]
 \centering
\begingroup
 \inputencoding{latin1} 
\includestandalone[width=0.48\textwidth]{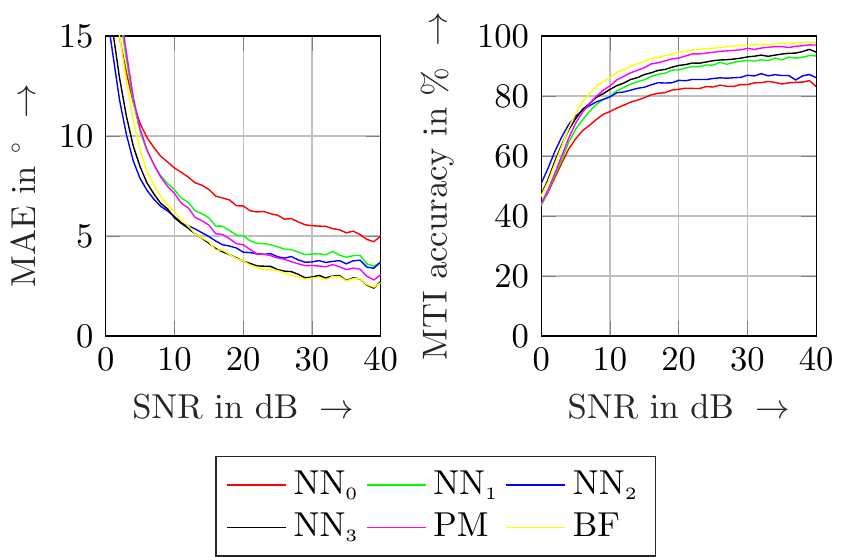}
\endgroup
    \caption{\textbf{Performance of DoA estimation methods:} 
Mean Absolute Error of azimuth estimation of different DoA estimators and performance of moving target indication (MTI).}
\label{fig:qualityCurveDoA}
\end{figure}

\inblue{
The performance of different DoA estimators in terms of \gls{MAE} is shown in Fig. \ref{fig:qualityCurveDoA}. In addition to \gls{NN}-based estimators we include the performance of a \gls{PM} \cite{Stabilito61} DoA estimator and \gls{BF} \cite{Krim96} as baseline algorithms which do not require a training phase.

As expected, it can be seen, that \gls{BF} achieved better MAE than \gls{PM} over the entire SNR range. 

The (3x3) \glspl{NN} achieved better MAE than their (1x1) counterparts, which suggests, that neighboring RD-map pixels carry valuable information that can be used for a better DoA estimation. A possible explanation for this is (a) the windowing performed before computing the FFT based RD grid, which smears energy over multiple pixels, and (b) the fact that geometrically neighboring pixels in the camera image tend to occupy neighboring locations in the RD-grid 

All \glspl{NN} with enabled SNR selection during training achieved significantly better MAE at higher SNR values. 

Finally it can be seen, that the best performing \gls{NN} ($\text{NN}_3$) achieved equal or better MAE accuracy as the best performing classical \gls{DOA} estimator (\gls{BF}).

}


Additionally we perform \gls{MTI}, which heavily relies on DoA predictions to classify stationary versus moving targets. The utilized MTI algorithm is from \cite{Grimm17a, Grimm17b}. If  $\|\boldsymbol{\xi}_{fg} \|_2^1 > \SI{0.5}{\meter\per\second}$ we set the reference label to ``moving'' and otherwise to ``stationary''. \gls{MTI} accuracy is calculated by comparing the prediction to the reference label. It can be observed in the figure that in low SNR regions, the (3x3) NN achieved better accuracy compared to the other estimators, while it is surpassed by beamforming in the high SNR regime. 

\inblue{
For a more detailed analysis, Fig. \ref{fig:qualityCurveDoA_hist} depicts the MAE as a function of the azimuth angle and SNR for selected estimators. Comparing the MAE histogram of $\text{NN}_0$ with $\text{NN}_1$  and $\text{NN}_2$ with $\text{NN}_3$, we see that the \glspl{NN} with enabled SNR selection ($\text{NN}_1$ and $\text{NN}_3$) achieved better MAE values especially at the bordering azimuth regions. This suggests, that optimizing the \gls{NN} on pixels from the signal mixture only, see \mbox{Fig. \ref{fig:snrDistribution_0}}, in combination with the used loss function, see \mbox{Eq. \ref{eq:lossFunction_DoA_0}}, is beneficial. However, to get isotropic DoA estimation behaviour over the entire azimuth range, it might be necessary to experiment with other loss functions in the future.

\begin{figure}[h]
 \centering
\begingroup
 \inputencoding{latin1} 
\includestandalone[width=\columnwidth]{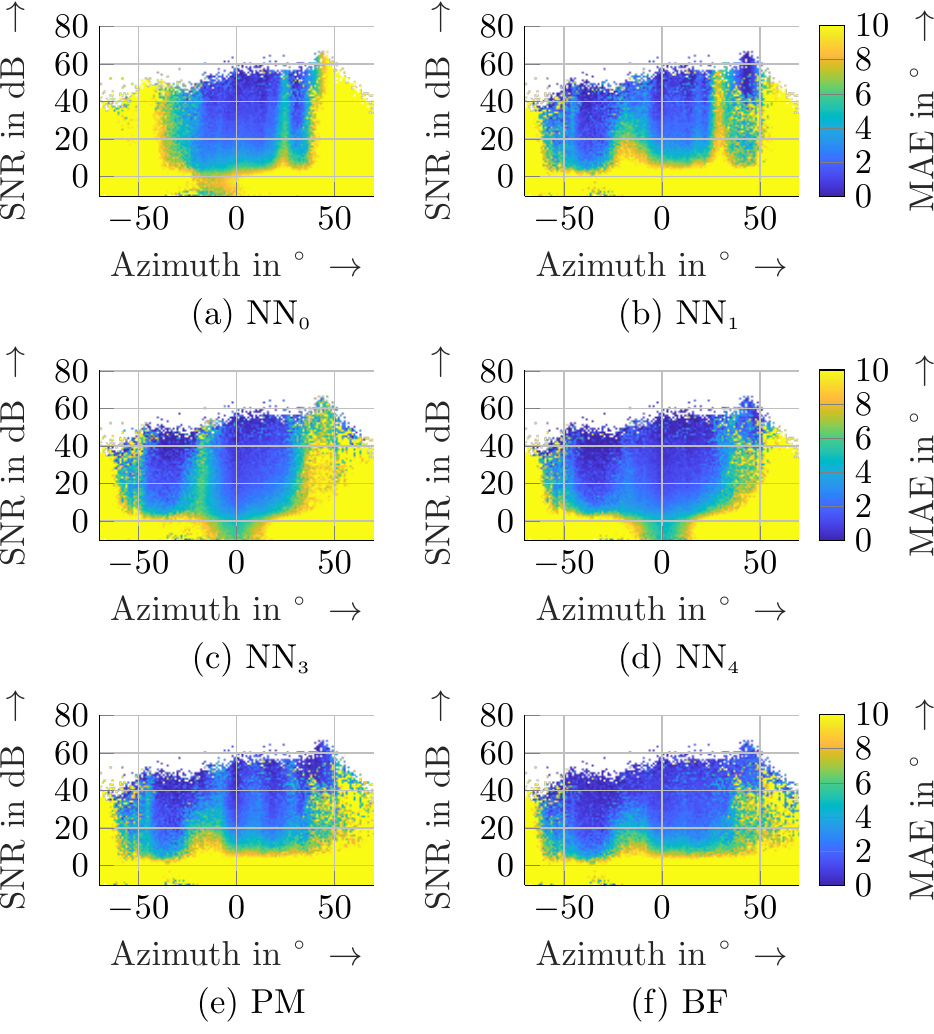}
\endgroup
\caption{\textbf{MAE histograms:} Mean Absolute Error in dependency on azimuth and SNR for different \gls{DOA} estimators. The \glspl{NN} with enabled SNR selection ($\text{NN}_1$ \& $\text{NN}_3$) achieved better MAE at bordering azimuth regions than their counterparts ($\text{NN}_0$ \& $\text{NN}_2$).}
\label{fig:qualityCurveDoA_hist}
\end{figure}

}

\begin{figure*}[ht]
 \centering
\begingroup
 \inputencoding{latin1} 
\includestandalone[width=\textwidth]{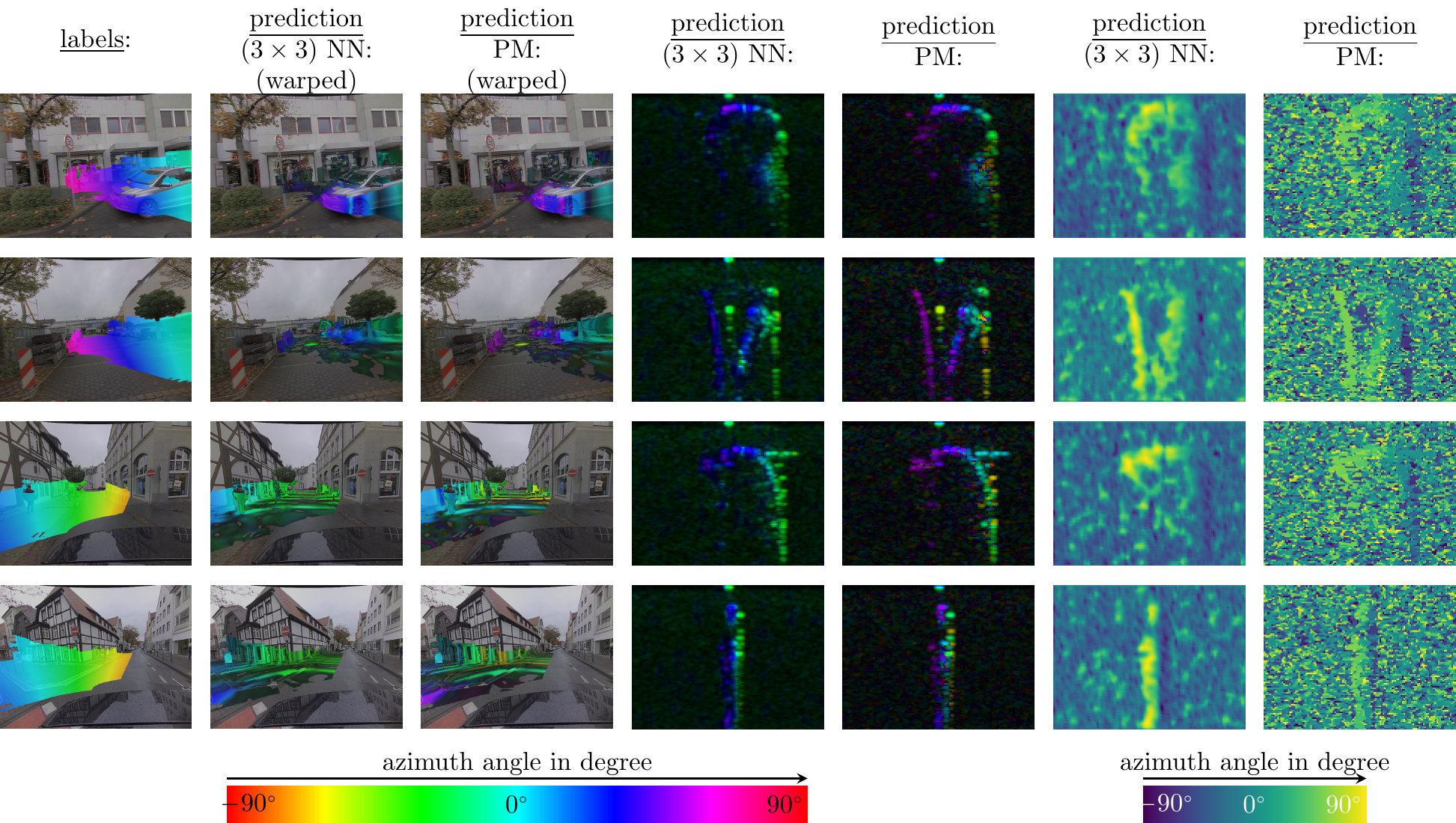}
\endgroup
    \caption{\textbf{Qualitative results of direction-of-arrival estimation on test data examples}. 
    From left-to-right: RGB image with color coded azimuth labels, RGB image with color coded azimuth predictions from NN, RGB image with color coded azimuth predictions from PM, RD-map with color coded azimuth prediction from NN, RD-map with color coded azimuth prediction from PM, azimuth predictions from NN in RD-grid, azimuth predictions from PM in RD-grid.}
    \label{fig:azimuthAngle2}
\end{figure*}

Qualitative examples for DoA estimation and \gls{MTI} are depicted in Fig. \ref{fig:azimuthAngle2}. It can be seen, that the color hue between DoA label and DoA prediction images mostly match and thus the predictions mostly resemble the DoA labels. The main difference between DoA label and prediction images comes from the selected visualization method, in which the color intensity of the prediction is scaled by the underlying RD-map SNR value. This visualization method helps to focus on regions with high SNR and reduces the distraction by low SNR regions, which mostly correspond to irrelevant noise regions. When comparing the non SNR color coded predictions from NN and PM (right hand side images in the Figure) for instance, it is obvious, that the NN predictions appear to be less noisy in low SNR regions. However, as stated above, DOA predictions in low SNR regions are in general of little interest.

As our framework provides full label density for a  camera image with roughly $185.000$ pixels per single frame, the DoA NN starts to provide reasonable estimations even after only 10 training steps/images. As the dataset consists of 72000 frames, we found this to be a very efficient usage of data.


\section{Conclusions}\label{ss:outlook}
In this paper we developed an algorithm for warping \gls{FMCW} radar data into camera image, enabling us to utilize high-quality labels generated from camera and lidar reference sensors for the supervised training of \glspl{NN} operating on radar spectra as input. An important factor for a reliable warping operation is the development of a novel scene flow estimation algorithm utilizing radar, camera and lidar data that is proposed in this paper.
The automatic labeling method  enabled us to densely label high-resolution spectrum level radar data from real world driving scenarios.

The validity of the approach was verified by evaluating the performance of  \glspl{NN} for DoA estimation, which achieved comparable or superior performance to classical DoA estimation methods. 
As the warping operation projects the spectrum level radar data and NN predictions into camera image, the results are easy to interpret for humans and allow qualitative assessment.

Furthermore, we could show that the use of the pixel vicinity through the NNs receptive field when performing DoA inference was beneficial. Although the general approach of utilizing pixel vicinity for better estimation is known from, e.g., classical image processing, to the best of our knowledge, we are the first to utilize this in radar based DoA  estimation. The development of this automatic labeling framework was key contribution for this observation,  



\bibliographystyle{IEEEtran}
\bibliography{IEEEabrv,literatur}


\begin{IEEEbiography}[{\includegraphics[width=1in,height=1.25in,clip,keepaspectratio]{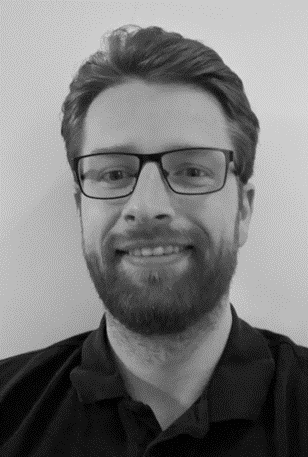}}]{Christopher Grimm}
(Graduate Student Member, IEEE) received the B. Eng. degree in automotive engineering from University of Applied Science and Arts Dortmund, Germany, in 2013, and the M. Sc. in mechanical engineering from Technical University of Braunschweig, Germany, in 2015. Since 2016 he is pursuing the Dr.-Ing. degree in electrical engineering with the Department of Communications Engineering, University of Paderborn, Germany and HELLA GmbH \& Co. KGaA, Lippstadt, Germany. Since 2020, he is working as a development engineer at HELLA GmbH \& Co. KGaA where he is responsible for the development of artificial intelligence based radar signal processing algorithms. His research interests are automotive radar signal processing, machine learning and computer vision.
\end{IEEEbiography}

\begin{IEEEbiography}[{\includegraphics[width=1in,height=1.25in,clip,keepaspectratio]{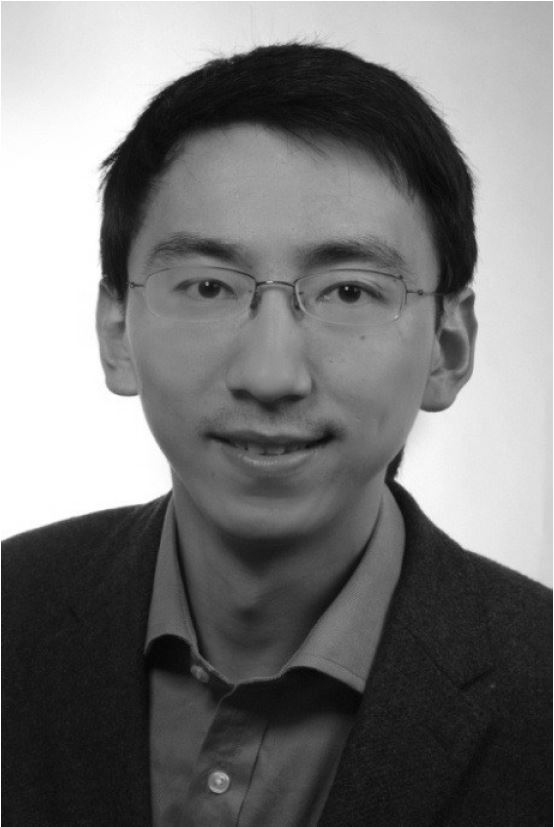}}]{Tai Fei}
(Senior Member, IEEE) received the B.Eng. degree in telecommunication engineering from Shanghai Maritime University, Shanghai, China, in 2005 and the Dipl.-Ing. and Dr.-Ing. degrees in electrical engineering and information technology from Darmstadt University of Technology (TUD), Darmstadt, Germany, in 2009 and 2014, respectively. From 2009 to 2012, he worked as a Research Associate with
the Institute of Water-Acoustics, Sonar-Engineering and Signal-Theory at Hochschule Bremen, Bremen,
Germany, in collaboration with the Signal Processing Group at TUD, Darmstadt, Germany, where his research interest was the detection and classification of underwater mines in sonar imagery. From
2013 to 2014, he worked as a Research Associate with the Center for Marine Environmental Sciences at University of Bremen, Bremen, Germany. Since 2014, has been working as a development engineer at HELLA GmbH \& Co. KGaA, Lippstadt, Germany, where he is mainly responsible for the development of reliable signal processing algorithms for automotive radar systems.
\end{IEEEbiography}

\begin{IEEEbiography}[{\includegraphics[width=1in,height=1.25in,clip,keepaspectratio]{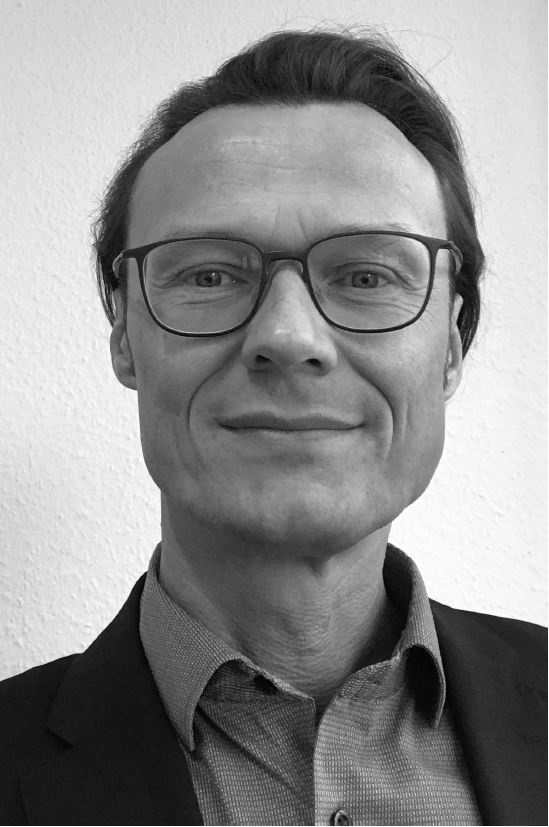}}]{Ernst Warsitz}
received the Dipl.-Ing. and Dr.-Ing. degrees in electrical engineering from Paderborn University, Paderborn, Germany, in 2000 and 2008, respectively. He joined the Department of Communications Engineering of the University of Paderborn in 2001 as a Research Staff Member, where he was involved in several projects relating to single- and multi-channel speech processing and
automated speech recognition. From 2007 he worked as a development engineer at HELLA GmbH \& Co.
KGaA, Lippstadt, Germany, in the field of signal processing algorithms for automotive radar systems. He is currently the head of the Radar Technology Department at HELLA GmbH \& Co. KGaA, Lippstadt, Germany.
\end{IEEEbiography}

\vfill
\newpage

\begin{IEEEbiography}[{\includegraphics[width=1in,height=1.25in,clip,keepaspectratio]{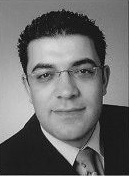}}]{Ridha Farhoud}
Ridha Farhoud received the Dipl.–Ing. degree in electrical engineering and the Dr.–Ing. degree from the Leipzig University in Hannover in 1997 and 2009, respectively. From 1998 to 2010, he worked as research engineer at the \glqq Institut für Informationsverarbeitung\grqq ~(information processing) in Hannover. 
He was involved in several projects related to Synthetic Aperture Radar, Image Interpretation and low bit rate video coding.
In 2010, he joined Hella KGaA Hueck \& Co. He is currently a member of the head of the Radar Signal Processing Department at HELLA GmbH \& Co. KGaA, Lippstadt, Germany.
\end{IEEEbiography}

\begin{IEEEbiography}[{\includegraphics[width=1in,height=1.25in,clip,keepaspectratio]{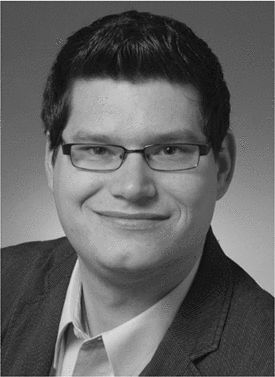}}]{Tobias Breddermann}
Tobias Breddermann received the Dipl.–Ing. degree in electrical engineering and the Dr.–Ing. degree from RWTH Aachen University, Aachen, Germany, in 2007 and 2013, respectively. From 2007 to 2013, he was with the Institute of Communication Systems and Data Processing at RWTH Aachen University. In 2013, he joined HELLA GmbH \& Co. KGaA. His research interests include radar singal processing, advanced DoA estimation and target separation.
\end{IEEEbiography}

\begin{IEEEbiography}[{\includegraphics[width=1in,height=1.25in,clip,keepaspectratio]{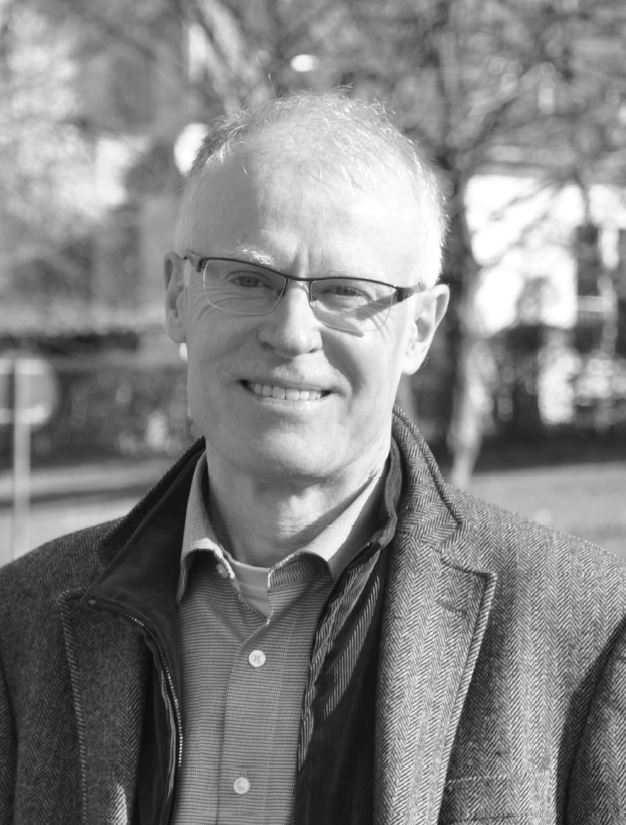}}]{Reinhold Haeb-Umbach}
(Fellow, IEEE) is a professor of Communications Engineering at Paderborn University, Germany. He holds a Dr.-Ing. degree from RWTH Aachen University, and has a background  both in an industrial and academic research environment. His main research interests are in the fields of statistical signal processing and machine learning, mainly with applications to speech and audio processing, but also to other fields, such as radar signal processing. He has authored more than 200 scientific publications, and he is both a fellow of the IEEE and of the  International Speech Communication Association (ISCA).
\end{IEEEbiography}

\vfill

\balance


\end{document}